\documentclass[runningheads]{llncs}

\usepackage{soul, color}
\usepackage[dvipsnames]{xcolor}
\usepackage[export]{adjustbox}  
\usepackage{fontawesome}
\usepackage{comment}
\usepackage{times}
\usepackage{epsfig}
\usepackage{amsmath}
\usepackage{cite}

\usepackage{bm}
\usepackage{mathtools}
\usepackage{amssymb}

\usepackage{float}

\usepackage{booktabs}
\usepackage{threeparttable}

\usepackage{graphicx}
\usepackage[lofdepth,lotdepth]{subfig}

\definecolor{myyellow}{RGB}{255, 255, 175}

\definecolor{mygreen}{RGB}{183, 255, 175}

\definecolor{myblue}{RGB}{208, 232, 255}

\definecolor{lyellow}{RGB}{255, 241, 180}

\newcommand{\m}[1]{\mathrm{#1}}
\newcommand{\mb}[1]{\mathbf{#1}}
\newcommand{\mx}{\mathbf{x}}

\newcommand{\mz}{\mb{z}}
\newcommand{\f}{\mb{f}}
\newcommand{\g}{\mb{g}}

\newcommand{\zB}[1]{\mathbf{z}_{\mathrm{b}}}
\newcommand{\zA}[1]{\mathbf{z}_{\mathrm{a}}}
\newcommand{\xA}[1]{\mathbf{x}_{a}^{#1}}
\newcommand{\xB}[1]{\mathbf{x}_{b}^{#1}}

\newcommand{\xa}{\mathbf{x}_{a}}
\newcommand{\xb}{\mathbf{x}_{b}}

\newcommand{\W}{\mathbf{W}}
\newcommand{\LL}{\mathbf{L}} 
\newcommand{\U}{\mathbf{U}}
\newcommand{\s}{\mathbf{s}}
\newcommand{\PP}{\mathbf{P}}

\renewcommand{\mid}{\,\ifnum\currentgrouptype=16 \middle\fi|\,}

\newcommand{\hk}[1]{{\color{black} #1}} 
\newcommand{\ms}[1]{{\color{black} #1}} 
\newcommand{\gh}[1]{{\color{black} #1}}

\newcommand{\FigScaleTemp}{0.33}

\newcommand{\FigScaleComp}{0.30}
\newcommand{\FigScaleContent}{0.33}
\newcommand{\DiverseGTScale}{.042}
\newcommand{\DiverseSampleScale}{0.084}
\newcommand{\GTLargeScale}{0.08}
\newcommand{\SingleSampleScale}{0.16}

\newcommand{\CompRowHSpace}{\hspace{0.3mm}}

\newcommand{\CompRowVSpace}{\vspace*{2mm}}

\newcommand{\TempText}{
    \mbox{$T=1.0$ \hspace{20mm} $T=0.8$ \hspace{20mm} $T=0.5$ \hspace{20mm} $T=0.0$}
}

\newcommand{\SegRealTextForTemp}{
    \mbox{\hspace{20mm} Conditioning \hspace{12mm} Ground-truth}
}

\newcommand{\SegRealText}{
    \mbox{\hspace{5mm} Conditioning \hspace{10mm} Ground-truth \hspace{6mm} Full-Glow sample 1 \hspace{1mm} Full-Glow sample 2}
}

\newcommand{\ModelNamesText}{
    \mbox{
    \hspace{0mm} C-Glow v.1 \cite{lu2020structured} \hspace{5mm} C-Glow v.2 \cite{lu2020structured} \hspace{5mm} Dual-Glow \cite{journals/corr/abs-1908-08074} \hspace{10mm} pix2pix \cite{Isola_2017_CVPR}}
}

\newcommand{\SingleCaption}{
    Higher resolution image synthesis by our model taken with temperature 0.9. Please zoom in to see more details.
}

\newcommand{\SingleText}{
    \mbox{\hspace{10mm} Conditioning \hspace{50mm} Synthesized}
}

\newcommand{\BlocksCaption}{\caption{Visual samples from different models. Samples from likelihood-based models are taken with temperature 0.7. Please zoom in to see more details.}}

\newcommand{\DiverseCaptionMultiple}{
    Higher resolution samples of our model taken with temperature 0.9. Left: conditioning, middle-right: samples 1, 2, 3. Please zoom in to see more details.
}

\newcommand{\ContentTransferText}{
    \mbox{\hspace{5mm} Desired content \hspace{8mm} Desired structure \hspace{5mm}  Synthesized image  \hspace{2mm} Ground-truth for structure}
}

\newcommand{\ContentTransferCaption}{
    \caption{Examples of applying a desired content to a desired structure. Please zoom in to see more details.}
}

\newcommand{\DrawTempFigTwo}{
    \begin{figure*}
    \centering
    \SegRealTextForTemp \newline
    \subfloat{\includegraphics[scale=\FigScaleTemp]{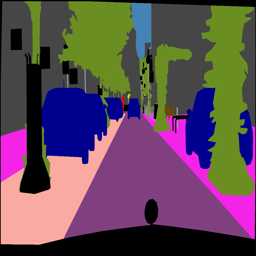}}\hspace{1mm}
    \subfloat{\includegraphics[scale=\FigScaleTemp]{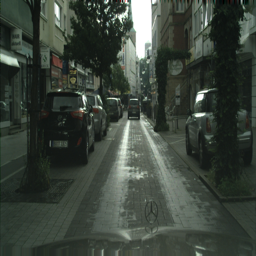}}
    \quad
    \TempText
    
    \subfloat{\includegraphics[scale=\FigScaleTemp]{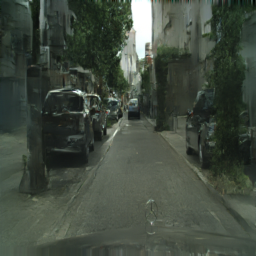}}
    \CompRowHSpace
    \subfloat{\includegraphics[scale=\FigScaleTemp]{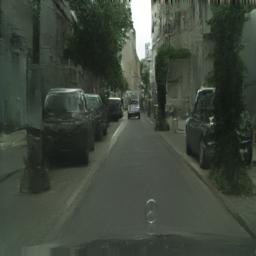}}
    \CompRowHSpace
    \subfloat{\includegraphics[scale=\FigScaleTemp]{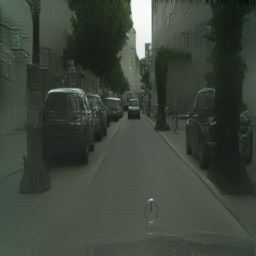}}
    \CompRowHSpace
    \subfloat{\includegraphics[scale=\FigScaleTemp]{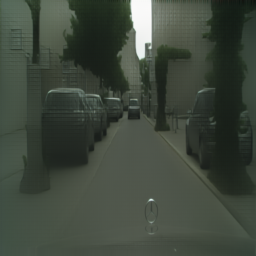}}
    \quad
    \subfloat{\includegraphics[scale=\FigScaleTemp]{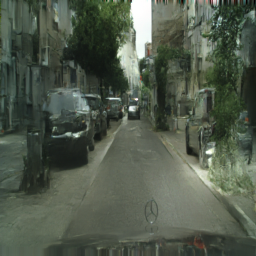}}
    \CompRowHSpace
    \subfloat{\includegraphics[scale=\FigScaleTemp]{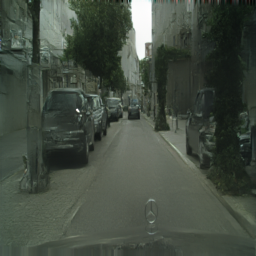}}
    \CompRowHSpace
    \subfloat{\includegraphics[scale=\FigScaleTemp]{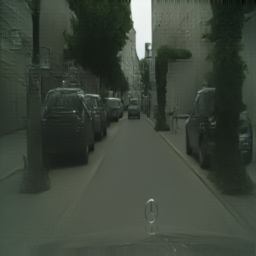}}
    \CompRowHSpace
    \subfloat{\includegraphics[scale=\FigScaleTemp]{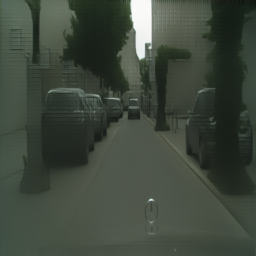}}
    \quad
    \subfloat{\includegraphics[scale=\FigScaleTemp]{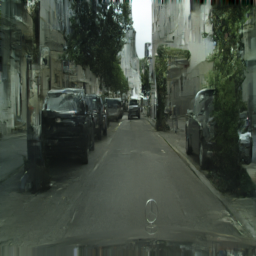}}
    \CompRowHSpace
    \subfloat{\includegraphics[scale=\FigScaleTemp]{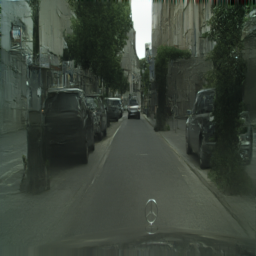}}
    \CompRowHSpace
    \subfloat{\includegraphics[scale=\FigScaleTemp]{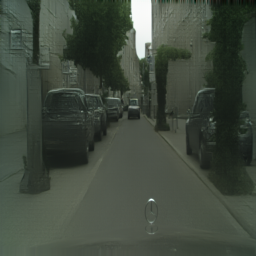}}
    \CompRowHSpace
    \subfloat{\includegraphics[scale=\FigScaleTemp]{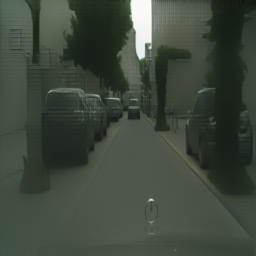}}
    \quad
    \caption{Effect of temperature: for each row, samples with temperature 1.0, 0.8, 0.5, and 0.0 are shown from left to right. Columns represent 3 samples with the given temperature. It is clear that samples with higher temperature show more diversity.}
    \label{fig:temp_effect_2}
    \end{figure*}
}

\newcommand{\DrawTempFigOne}{
    \begin{figure*}
    \centering
    \SegRealTextForTemp \newline
    \subfloat{\includegraphics[scale=\FigScaleTemp]{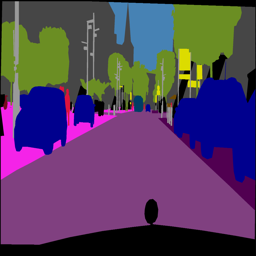}}\hspace{1mm}
    \subfloat{\includegraphics[scale=\FigScaleTemp]{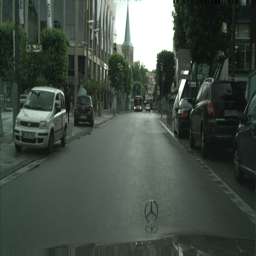}}
    \quad
    \TempText
    
    \subfloat{\includegraphics[scale=\FigScaleTemp]{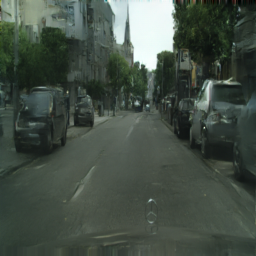}}
    \CompRowHSpace
    \subfloat{\includegraphics[scale=\FigScaleTemp]{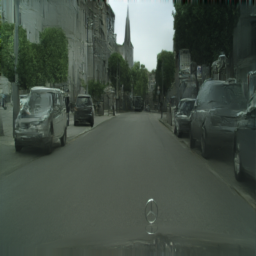}}
    \CompRowHSpace
    \subfloat{\includegraphics[scale=\FigScaleTemp]{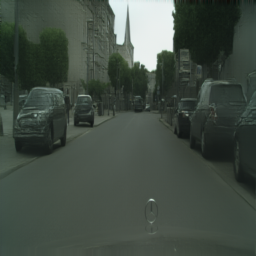}}
    \CompRowHSpace
    \subfloat{\includegraphics[scale=\FigScaleTemp]{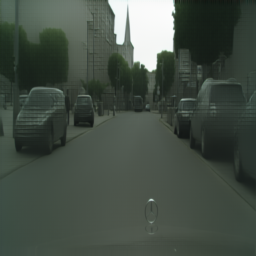}}
    \quad
    \subfloat{\includegraphics[scale=\FigScaleTemp]{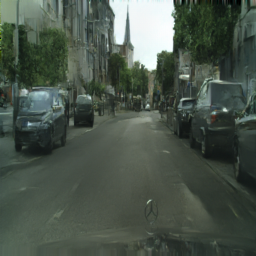}}
    \CompRowHSpace
    \subfloat{\includegraphics[scale=\FigScaleTemp]{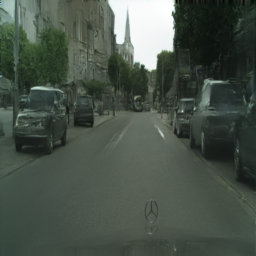}}
    \CompRowHSpace
    \subfloat{\includegraphics[scale=\FigScaleTemp]{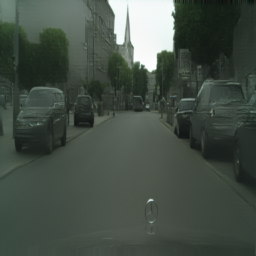}}
    \CompRowHSpace
    \subfloat{\includegraphics[scale=\FigScaleTemp]{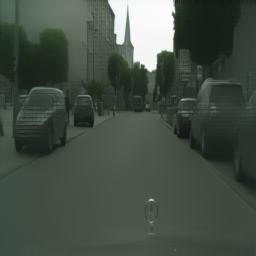}}
    \quad
    \subfloat{\includegraphics[scale=\FigScaleTemp]{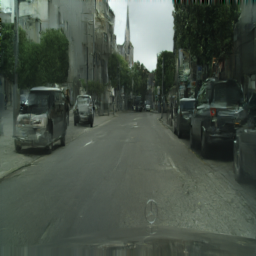}}
    \CompRowHSpace
    \subfloat{\includegraphics[scale=\FigScaleTemp]{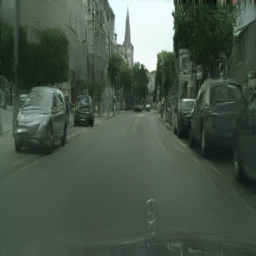}}
    \CompRowHSpace
    \subfloat{\includegraphics[scale=\FigScaleTemp]{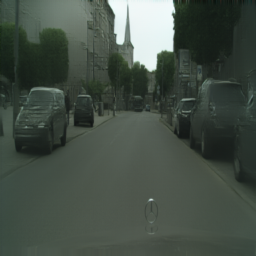}}
    \CompRowHSpace
    \subfloat{\includegraphics[scale=\FigScaleTemp]{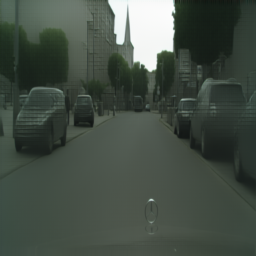}}
    \quad
    \caption{Effect of temperature: for each row, samples with temperature 1.0, 0.8, 0.5, and 0.0 are shown from left to right. Columns represent 3 samples with the given temperature. It is clear that samples with higher temperature show more diversity.}
    \label{fig:temp_effect_1}
    \end{figure*}
}

\newcommand{\DrawTempTable}{
    \begin{table}[]
        \centering
        \small
        \caption{Effect of temperature $T$ evaluated using a pre-trained PSPNet \cite{zhao2017pyramid}. \gh{Each column lists the mean over repeated image samples.}}
        \label{tab:temp_effect}
        \begin{tabular}{@{}l|lll@{}}
        $\boldsymbol{T}$ & \textbf{Pixel acc.} & \textbf{Class acc.} & \textbf{Class IoU}       \\
        \hline
        0.0         & 27.83 $\pm$ 0   & 12.48 $\pm$ 0   & \hphantom{0}7.69 $\pm$ 0   \\
        0.1         & 29.47 $\pm$ 0.04  & 13.15 $\pm$ 0.05  & \hphantom{0}8.33 $\pm$ 0.01  \\
        0.2         & 34.97 $\pm$ 0.13  & 15.83 $\pm$ 0.13  & 10.65 $\pm$ 0.01 \\
        0.3         & 43.36 $\pm$ 0.08  & 20.22 $\pm$ 0.05  & 14.69 $\pm$ 0.02 \\
        0.4         & 55.26 $\pm$ 0.12  & 25.90 $\pm$ 0.15   & 20.01 $\pm$ 0.11 \\
        0.5         & 69.83 $\pm$ 0.01  & 31.66 $\pm$ 0.16  & 25.82 $\pm$ 0.12 \\
        0.6         & 79.38 $\pm$ 0.05  & 34.90 $\pm$ 0.28   & 29.37 $\pm$ 0.23 \\
        0.7         & 82.14 $\pm$ 0.09  & 35.51 $\pm$ 0.23  & 29.98 $\pm$ 0.21 \\
        \textbf{0.8}         & \textbf{83.52 $\pm$ 0.04}  & \textbf{35.94 $\pm$ 0.19}  & \textbf{30.67 $\pm$ 0.17} \\
        0.9         & 82.27 $\pm$ 0.05  & 35.24 $\pm$ 0.37  & 29.85 $\pm$ 0.36 \\
        1.0         & 73.50 $\pm$ 0.13     & 29.13 $\pm$ 0.39    & 23.86 $\pm$ 0.30    
        \end{tabular}
    \end{table}
}

\newcommand{\DrawComparisonTable}{
    \begin{table}[!b]
        \centering
        \footnotesize
        \caption{Comparison of different models \gh{on} the Cityscapes dataset for label $\rightarrow$ photo image synthesis.}
        \label{tab:models_comparison}
        \begin{tabular}{@{}l|llll@{}}
        \textbf{Model}        & \textbf{\begin{tabular}[c]{@{}l@{}}Cond.\\ BPD\end{tabular}} & \textbf{\begin{tabular}[c]{@{}l@{}}Mean \\ pixel acc.\end{tabular}} & \textbf{\begin{tabular}[c]{@{}l@{}}Mean\\ class acc.\end{tabular}} & \textbf{\begin{tabular}[c]{@{}l@{}}Mean \\ class IoU\end{tabular}}    \\ \hline
        C-Glow v.1 \cite{lu2020structured}      & 2.568                    & 35.02 ± 0.56             & 12.15 ± 0.05             & \hphantom{0}7.33 ± 0.09          \\
        C-Glow v.2 \cite{lu2020structured}      & 2.363                    & 52.33 ± 0.46             & 17.37 ± 0.21             & 12.31 ± 0.24         \\
        Dual-Glow  \cite{journals/corr/abs-1908-08074}  & 2.585                    & 71.44 ± 0.03             & 23.91 ± 0.19             & 18.96 ± 0.17         \\
        pix2pix \cite{Isola_2017_CVPR}  & ---                      & 60.56 ± 0.11             & 22.64 ± 0.21             & 16.42 ± 0.06         \\
        \textbf{Our model}    & \textbf{2.345}           & \textbf{73.50 ± 0.13}     & \textbf{29.13 ± 0.39}    & \textbf{23.86 ± 0.30} \\ \hline
        \textit{Ground-truth} & \textit{---}             & \textit{95.97}           & \textit{84.31}           & \textit{77.30}       
        \end{tabular}
    \end{table}
}

\newcommand{\DrawBlockTwo}{
    \centering
    \SegRealText \newline
    \subfloat{\includegraphics[scale=\FigScaleComp]{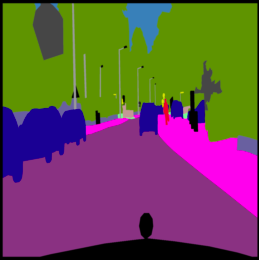}}
    \CompRowHSpace
    \subfloat{\includegraphics[scale=\FigScaleComp]{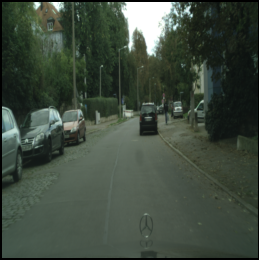}}
    \CompRowHSpace
    \subfloat{\includegraphics[scale=\FigScaleComp]{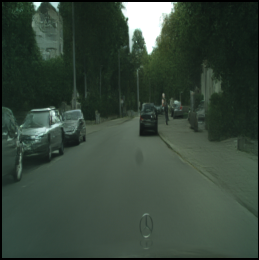}}
    \CompRowHSpace
    \subfloat{\includegraphics[scale=\FigScaleComp]{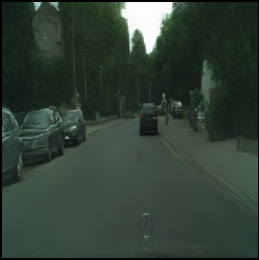}}
    \CompRowVSpace
    \ModelNamesText
    
    \subfloat{\includegraphics[scale=\FigScaleComp]{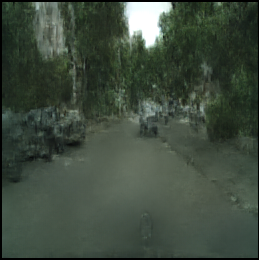}}
    \CompRowHSpace
    \subfloat{\includegraphics[scale=\FigScaleComp]{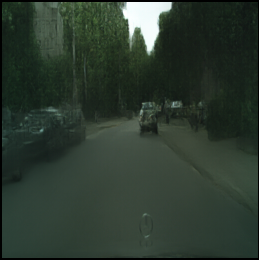}}
    \CompRowHSpace
    \subfloat{\includegraphics[scale=\FigScaleComp]{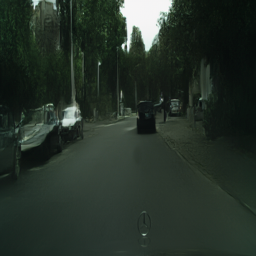}}
    \CompRowHSpace
    \subfloat{\includegraphics[scale=\FigScaleComp]{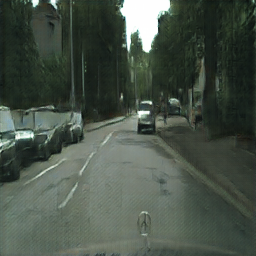}}
}

\newcommand{\DrawBlockThree}{
    \centering
    \SegRealText \newline
    \subfloat{\includegraphics[scale=\FigScaleComp]{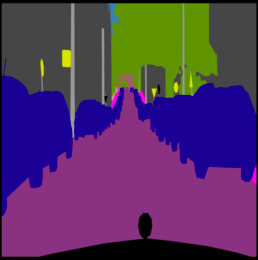}}
    \CompRowHSpace
    \subfloat{\includegraphics[scale=\FigScaleComp]{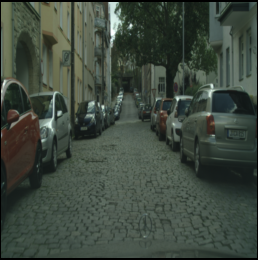}}
    \CompRowHSpace
    \subfloat{\includegraphics[scale=\FigScaleComp]{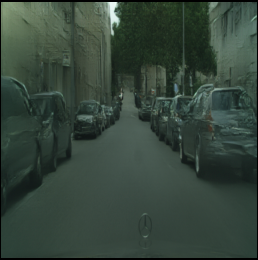}}
    \CompRowHSpace
    \subfloat{\includegraphics[scale=\FigScaleComp]{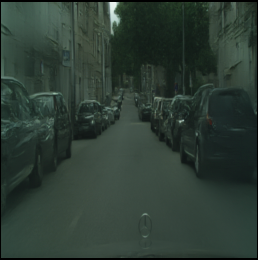}}
    \CompRowVSpace
    \ModelNamesText
    
    \subfloat{\includegraphics[scale=\FigScaleComp]{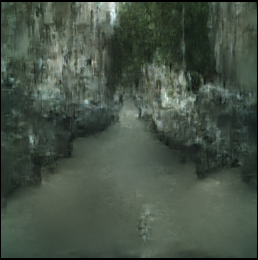}}
    \CompRowHSpace
    \subfloat{\includegraphics[scale=\FigScaleComp]{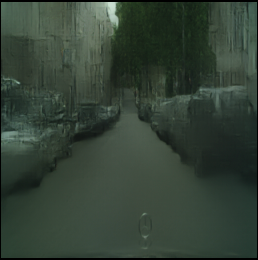}}
    \CompRowHSpace
    \subfloat{\includegraphics[scale=\FigScaleComp]{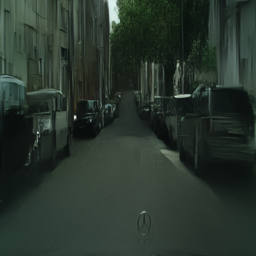}}
    \CompRowHSpace
    \subfloat{\includegraphics[scale=\FigScaleComp]{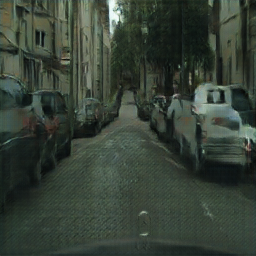}}
}

\newcommand{\DrawContentTransferOne}{
        \subfloat{\includegraphics[scale=\FigScaleContent]{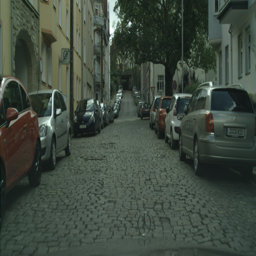}}
        \CompRowHSpace
        \subfloat{\includegraphics[scale=\FigScaleContent]{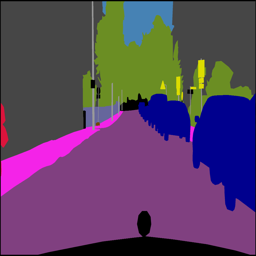}}
        \CompRowHSpace
        \subfloat{\includegraphics[scale=\FigScaleContent]{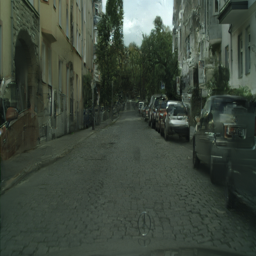}}
        \CompRowHSpace
        \subfloat{\includegraphics[scale=\FigScaleContent]{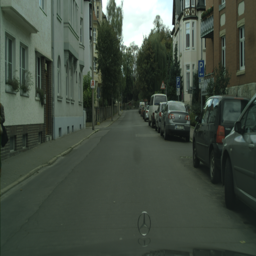}}
        
}

\newcommand{\DrawContentTransferTwo}{
        \subfloat{\includegraphics[scale=\FigScaleContent]{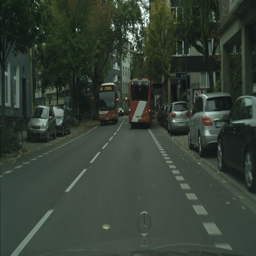}}
        \CompRowHSpace
        \subfloat{\includegraphics[scale=\FigScaleContent]{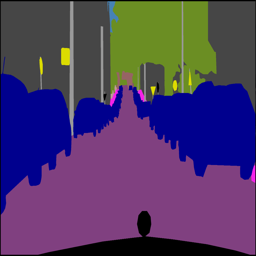}}
        \CompRowHSpace
        \subfloat{\includegraphics[scale=\FigScaleContent]{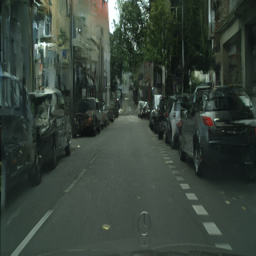}}
        \CompRowHSpace
        \subfloat{\includegraphics[scale=\FigScaleContent]{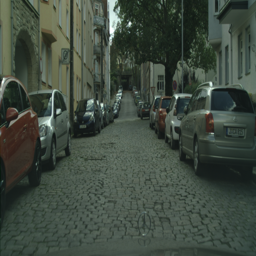}}
        
}

\newcommand{\DrawContentTransferThree}{
        \subfloat{\includegraphics[scale=\FigScaleContent]{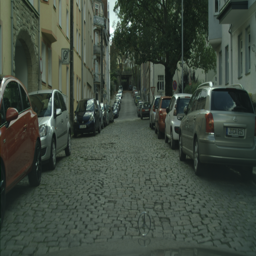}}
        \CompRowHSpace
        \subfloat{\includegraphics[scale=\FigScaleContent]{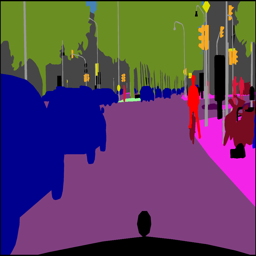}}
        \CompRowHSpace
        \subfloat{\includegraphics[scale=\FigScaleContent]{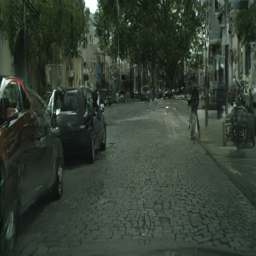}}
        \CompRowHSpace
        \subfloat{\includegraphics[scale=\FigScaleContent]{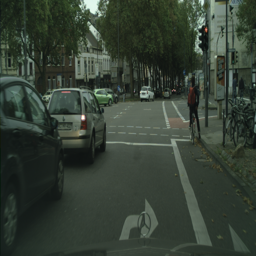}}
        
}

\newcommand{\DrawDeepTable}{
    \begin{table*}[ht!]
        \scriptsize
        \caption{Performance of different configurations of our model, showing the effect of using boundary maps and deepening the models (in 256$\times$256 resolution).}
        \label{tab:b_map_with_deep}
        \begin{tabular}{l|l|l|l|rrr}
        \textbf{Model}                                  & \textbf{Description}                     & \textbf{\begin{tabular}[c]{@{}l@{}}Total \\ Flows\end{tabular}} & \textbf{Params} & \multicolumn{1}{l}{\textbf{\begin{tabular}[c]{@{}l@{}}Mean \\ pixel acc.\end{tabular}}} & \multicolumn{1}{l}{\textbf{\begin{tabular}[c]{@{}l@{}}Mean \\ class acc.\end{tabular}}} & \multicolumn{1}{l}{\textbf{\begin{tabular}[c]{@{}l@{}}Mean \\ class IoU\end{tabular}}} \\ \hline
        Config. A - Baseline                            & 4 Blocks, 16 Flows                       & 64                                                              & 155.3M          & 73.50 ± 0.13                                                                             & 29.13 ± 0.39                                                                            & 23.86 ± 0.30                                                                            \\
        Config. B - Using b-maps v.1                    & 4 Blocks, 16 Flows                       & 64                                                              & 155.5M          & 74.95 ± 0.03                                                                            & 30.46 ± 0.27                                                                            & 25.24 ± 0.08                                                                           \\
        Config. C - Using b-maps v.2                    & 4 Blocks, 16 Flows                       & 64                                                              & 155.5M          & 74.12 ± 0.13                                                                            & 28.96 ± 0.04                                                                            & 24.14 ± 0.07                                                                           \\
        Config. D - Deep model                          & 5 Blocks, 24 Flows                       & 120                                                             & 431M            & 75.13 ± 0.28                                                                            & 29.71 ± 0.20                                                                             & 24.32 ± 0.30                                                                            \\
        Config. E - Deep hierarchical                   & 5 Blocks, {[}40, 30, 20, 20, 10{]} Flows & 120                                                             & 348M            & 77.38 ± 0.36                                                                            & 31.46 ± 0.43                                                                            & 26.30 ± 0.34                                                                            \\
        Config. F - Deep with fewer Blocks v.1          & 4 Blocks, 30 Flows                       & 120                                                             & 290M            & 74.97 ± 0.01                                                                            & 29.93 ± 0.28                                                                            & 25.07 ± 0.19                                                                           \\
        \textbf{Config. H - Deep with fewer Blocks v.2} & \textbf{3 Blocks, 40 Flows}              & \textbf{120}                                                    & \textbf{248M}   & \textbf{77.90 ± 0.14}                                                                    & \textbf{32.89 ± 0.35}                                                                   & \textbf{27.82 ± 0.27}                                                                 
        \end{tabular}
    \end{table*}
}

\newcommand{\DrawDiverseOne}{
    \subfloat{\includegraphics[scale=\DiverseGTScale]{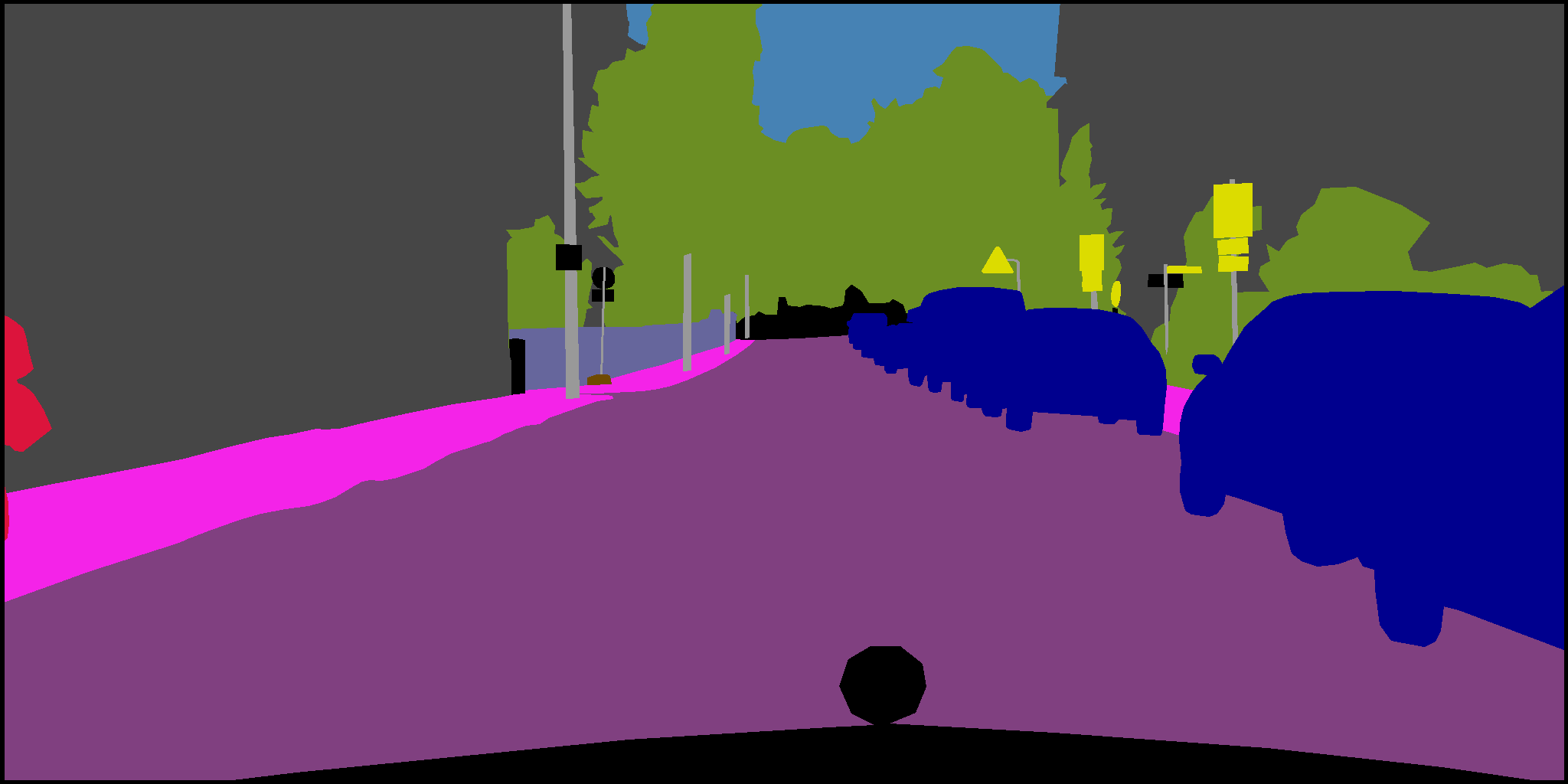}}
    \subfloat{\includegraphics[scale=\DiverseSampleScale]{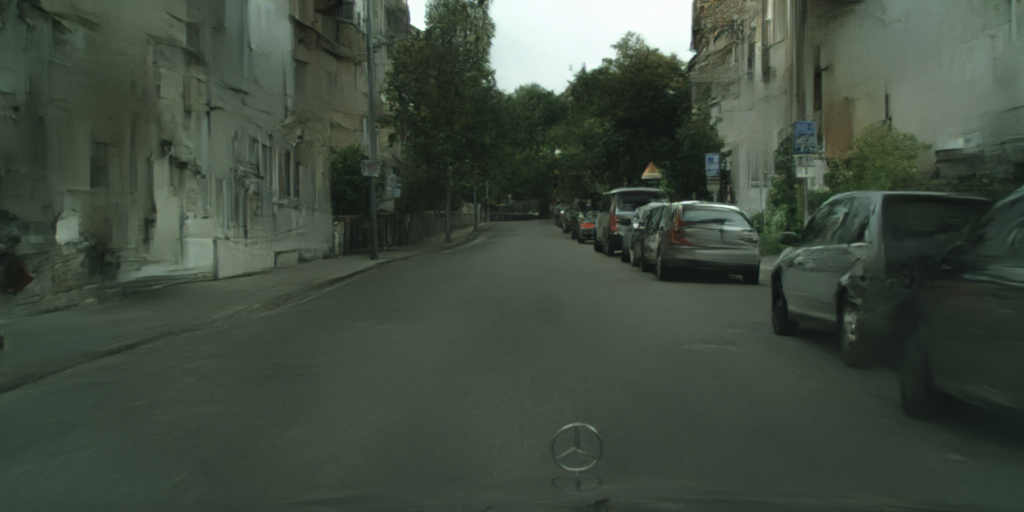}}
    \subfloat{\includegraphics[scale=\DiverseSampleScale]{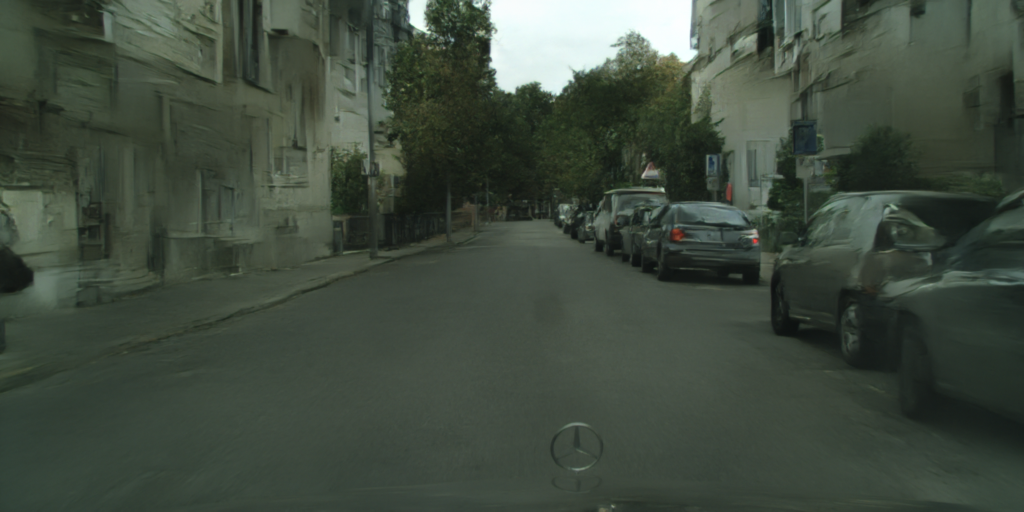}}
    \subfloat{\includegraphics[scale=\DiverseSampleScale]{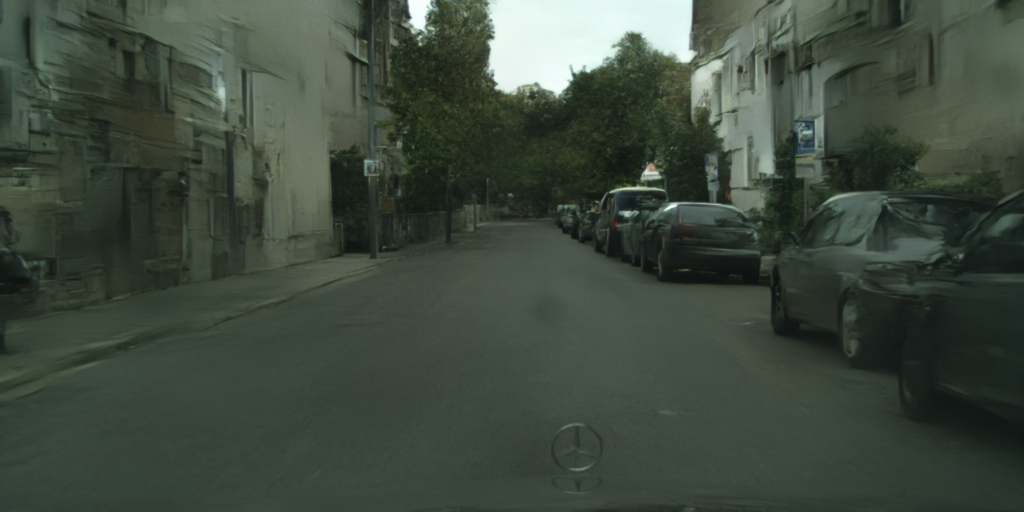}}
}

\newcommand{\DrawDiverseTwo}{
    \subfloat{\includegraphics[scale=\DiverseGTScale]{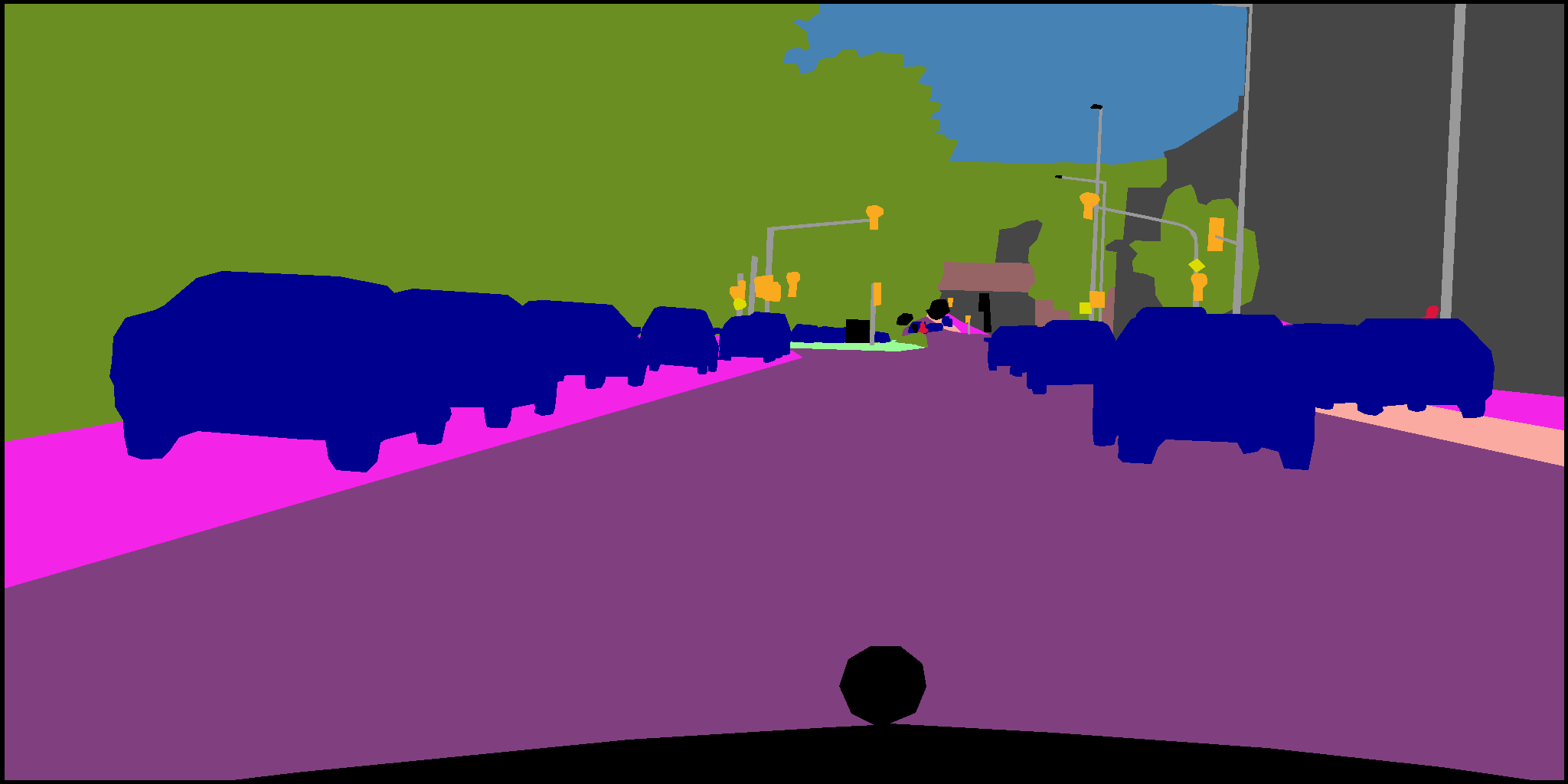}}
    \subfloat{\includegraphics[scale=\DiverseSampleScale]{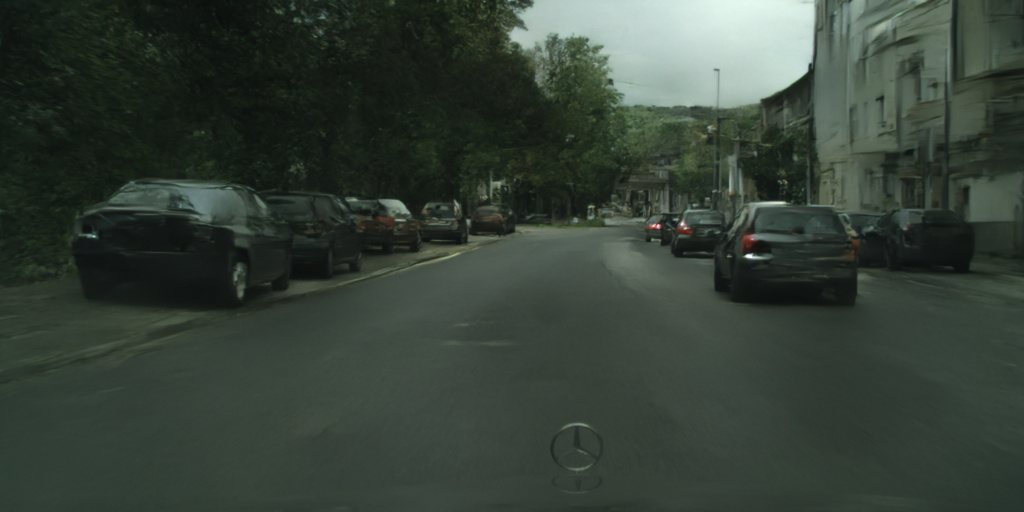}}
    \subfloat{\includegraphics[scale=\DiverseSampleScale]{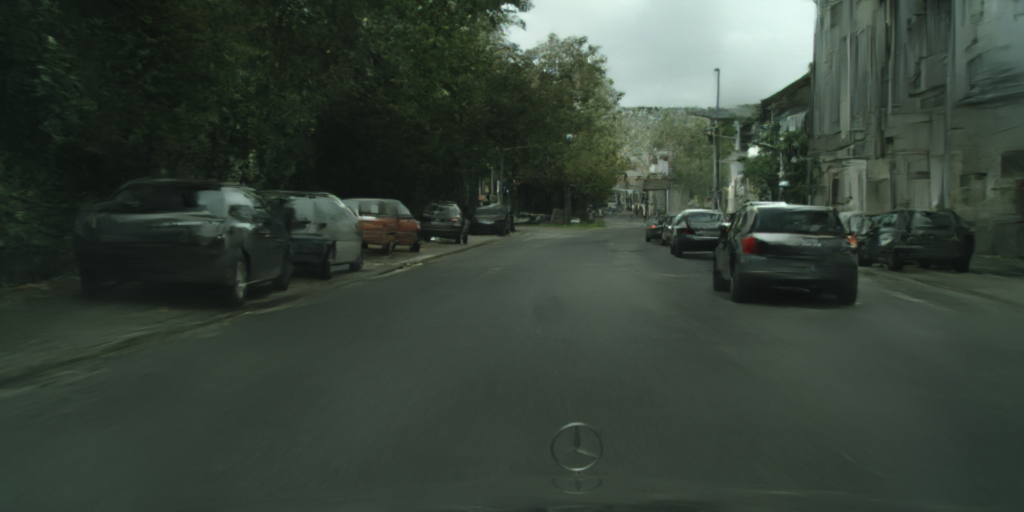}}
    \subfloat{\includegraphics[scale=\DiverseSampleScale]{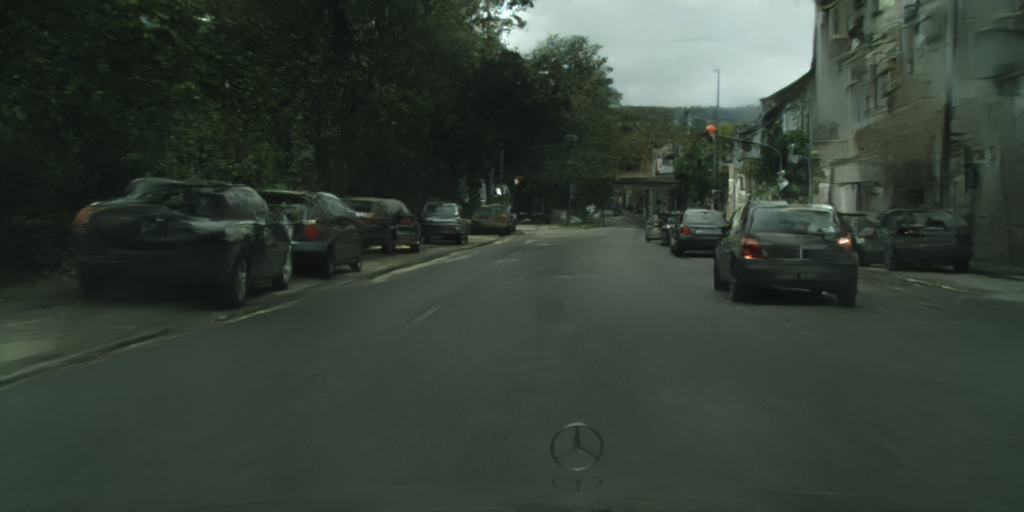}}
}

\newcommand{\DrwaSingleOne}{
    \SingleText \newline
    \subfloat{\includegraphics[scale=\GTLargeScale]{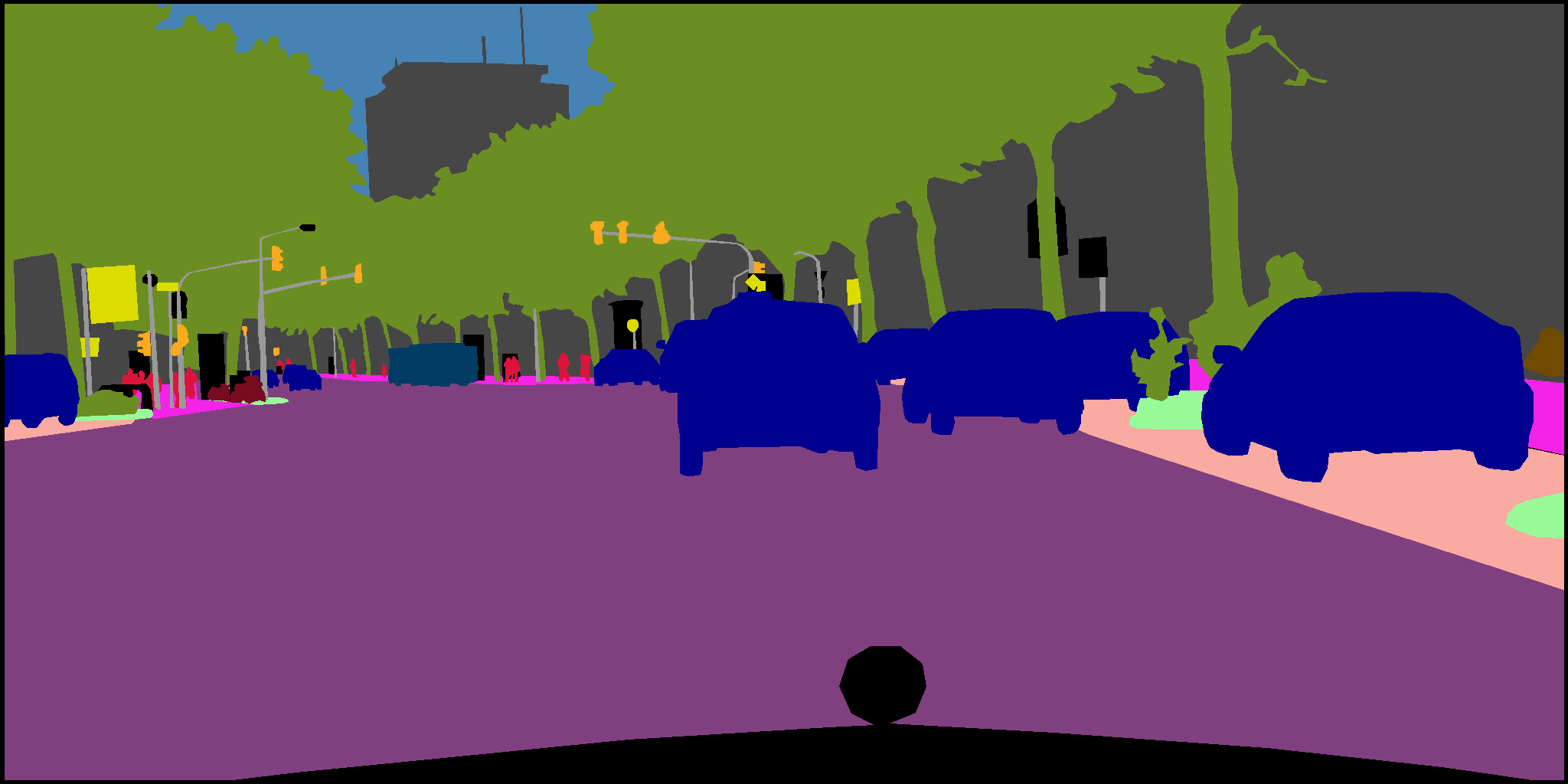}}
    \hfill
    \subfloat{\includegraphics[scale=\SingleSampleScale]{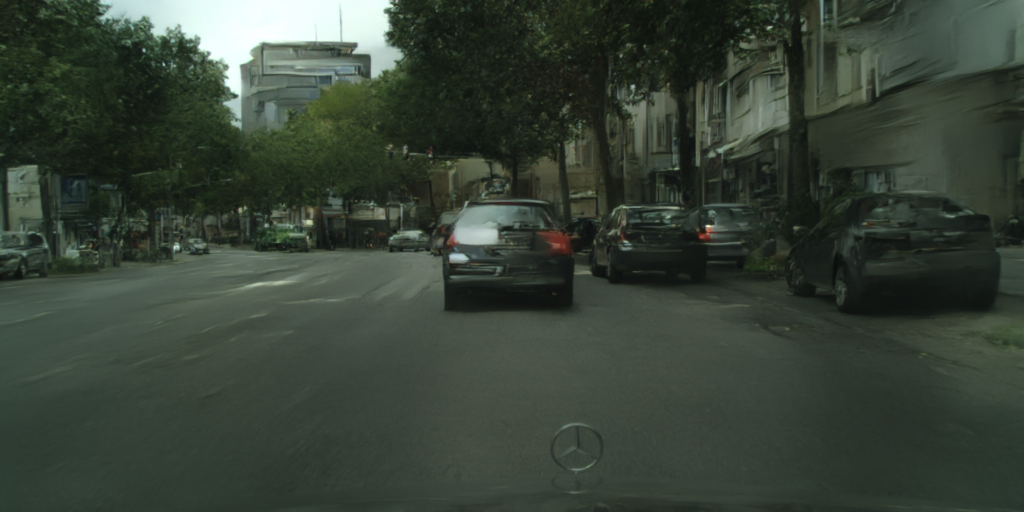}}
    \quad
    \subfloat{\includegraphics[scale=\GTLargeScale]{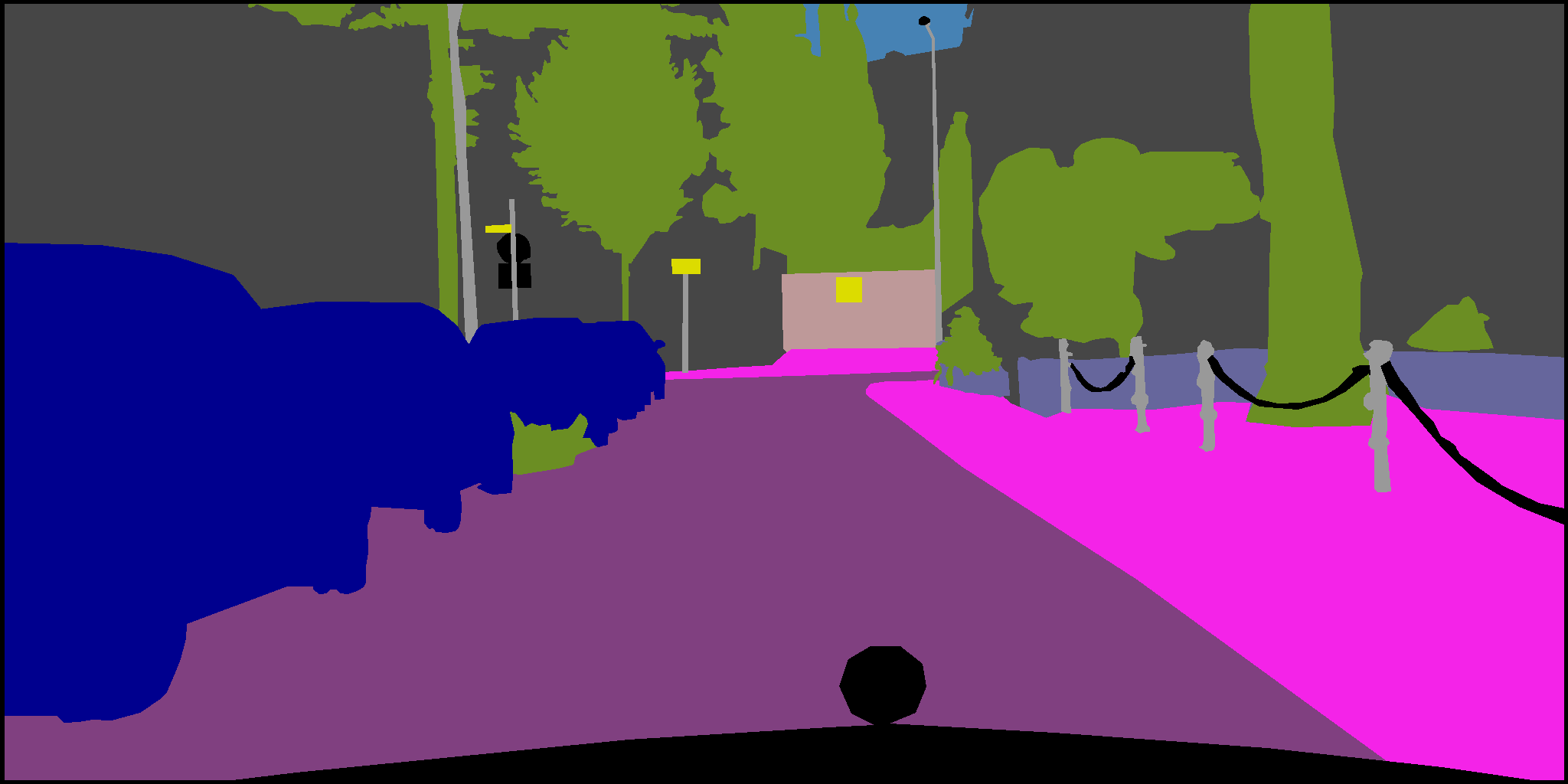}}
    \hfill
    \subfloat{\includegraphics[scale=\SingleSampleScale]{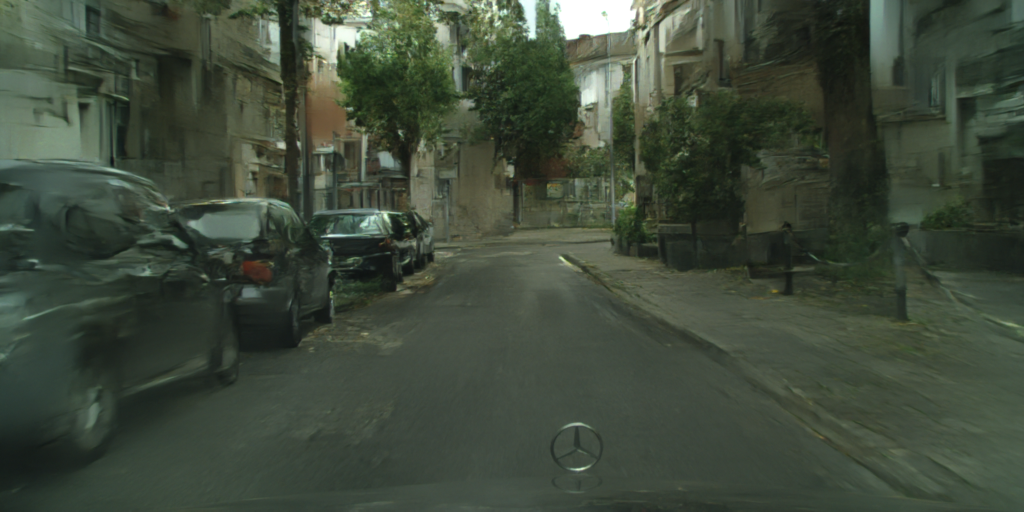}}
    \quad
    \subfloat{\includegraphics[scale=\GTLargeScale]{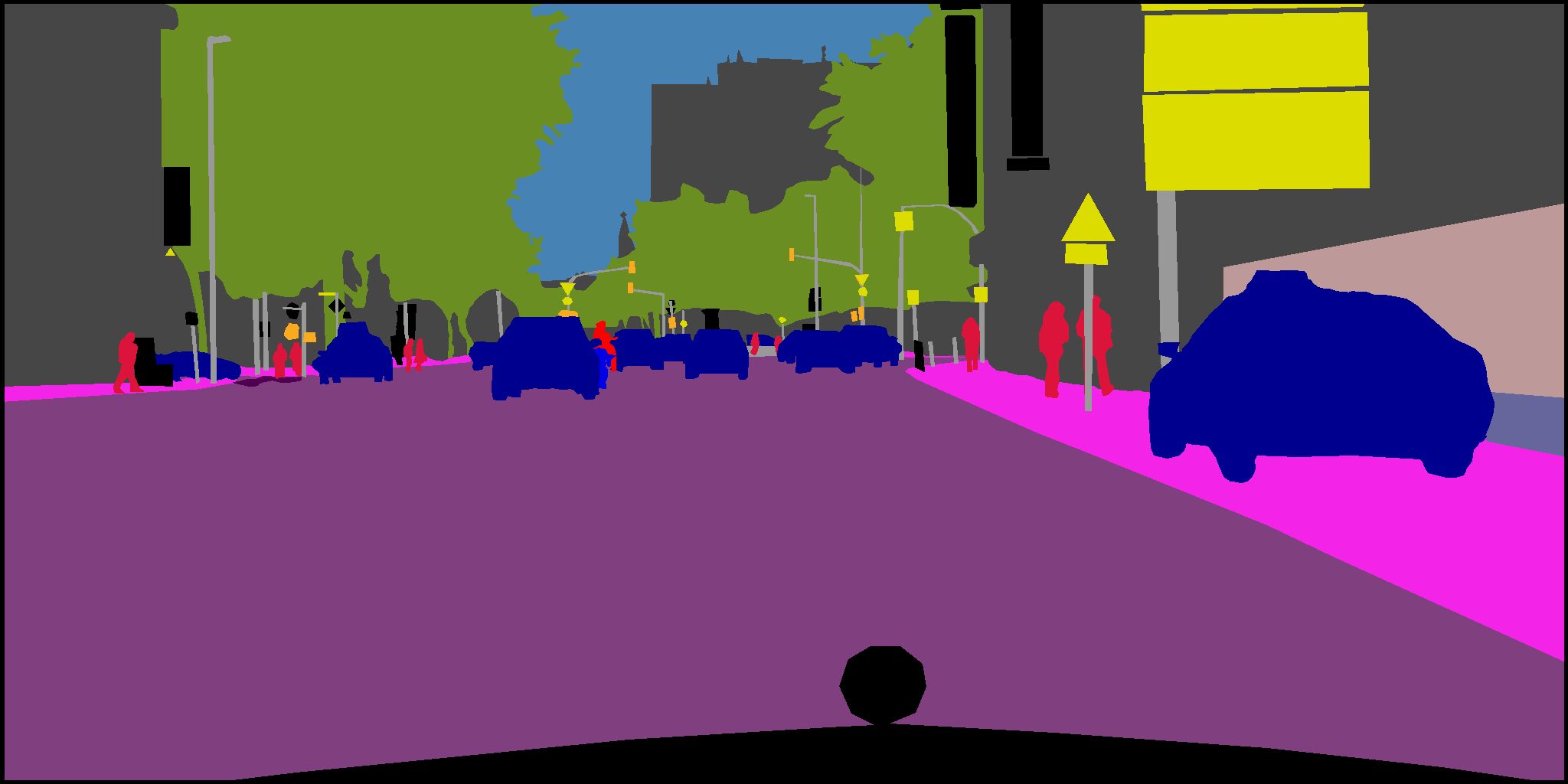}}
    \hfill
    \subfloat{\includegraphics[scale=\SingleSampleScale]{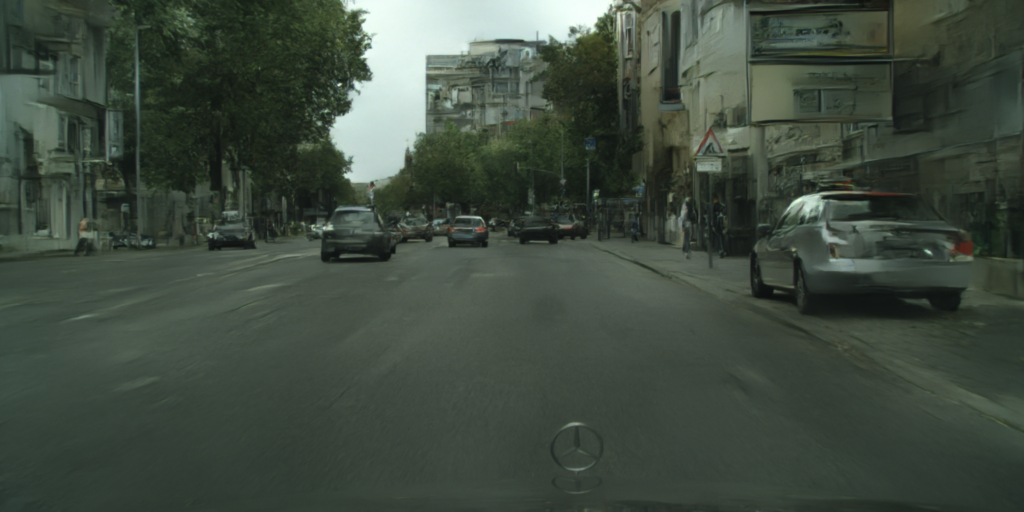}}
    \quad
    \subfloat{\includegraphics[scale=\GTLargeScale]{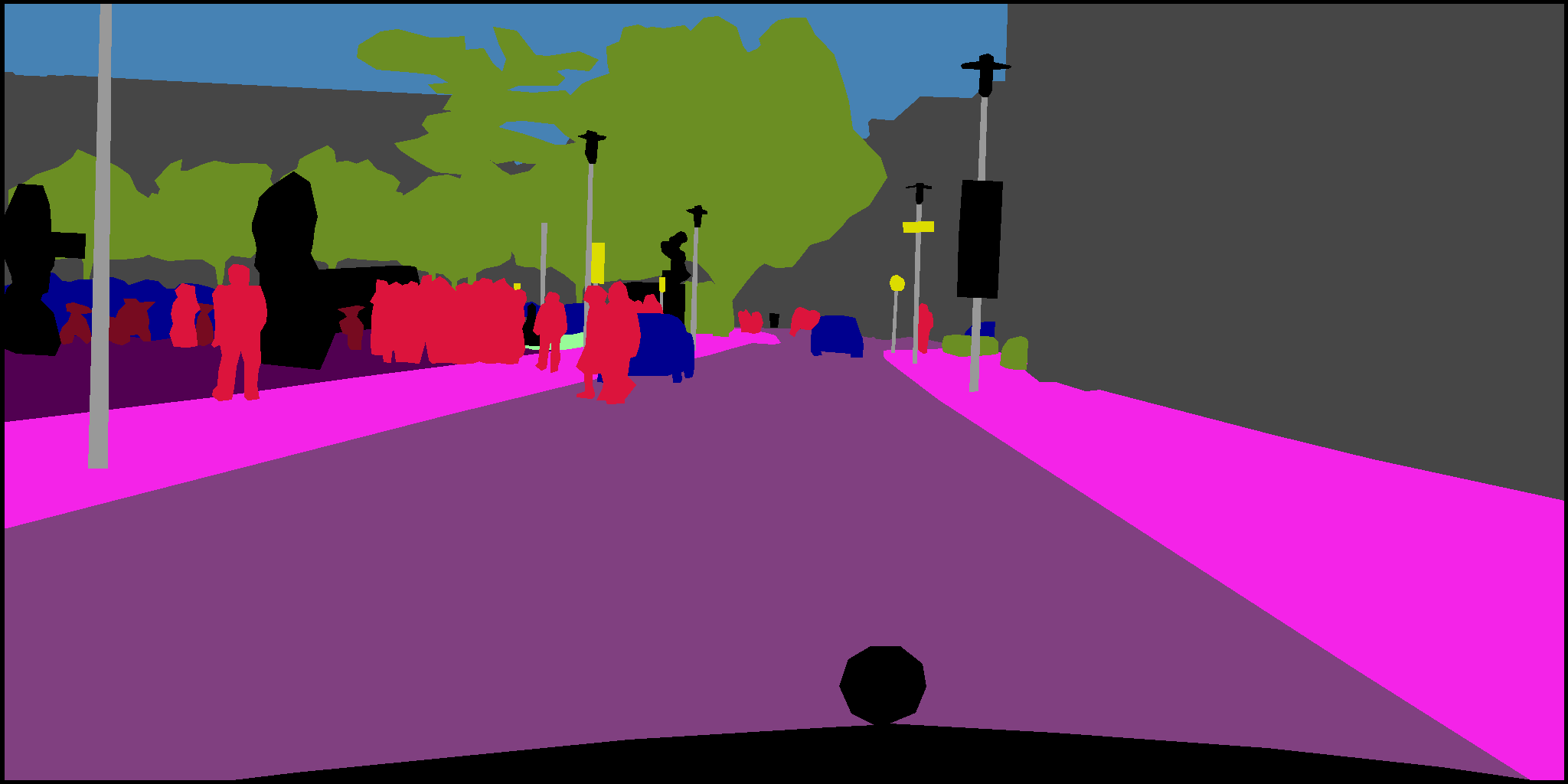}}
    \hfill
    \subfloat{\includegraphics[scale=\SingleSampleScale]{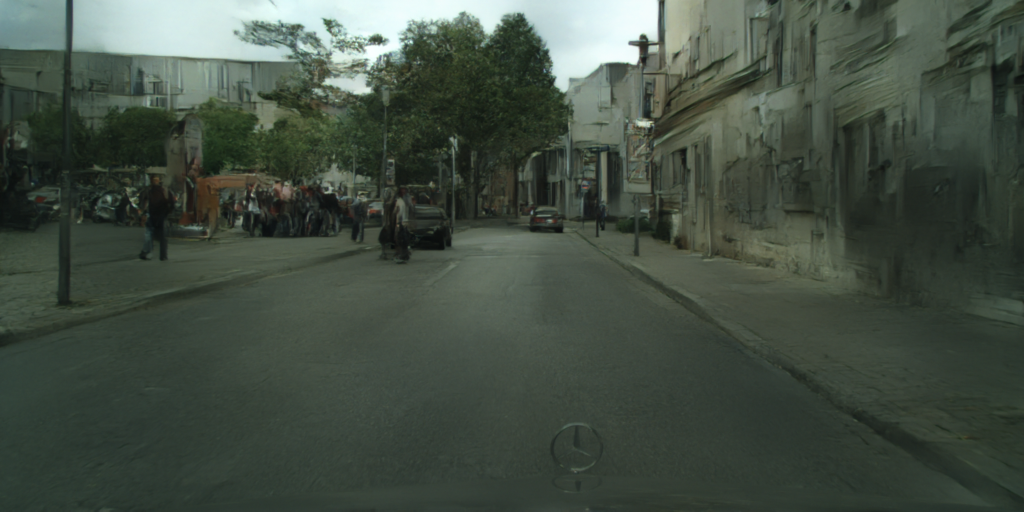}}
}

\newcommand{\DrwaSingleTwo}{
    \SingleText \newline
    \subfloat{\includegraphics[scale=\GTLargeScale]{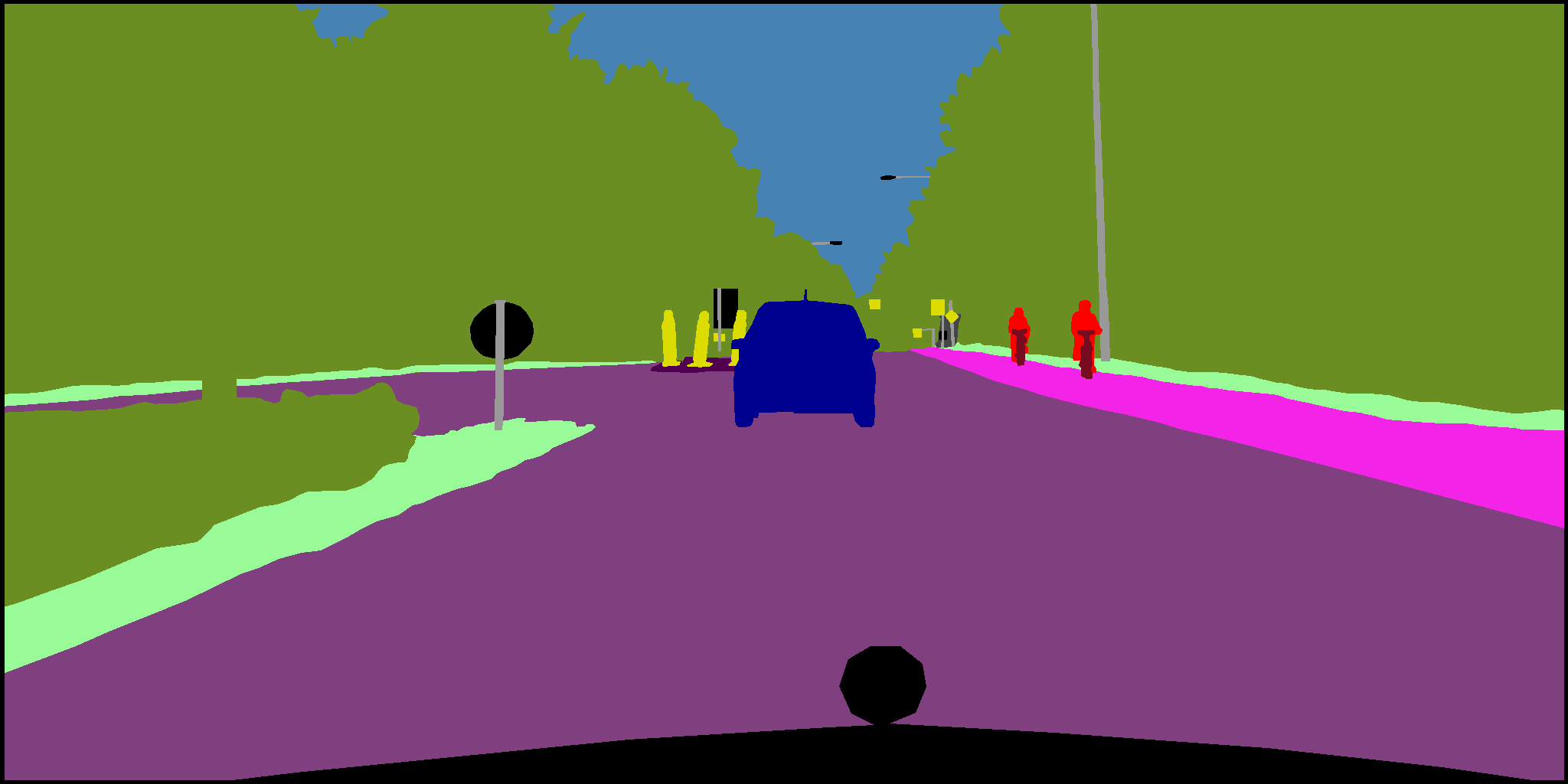}}
    \hfill
    \subfloat{\includegraphics[scale=\SingleSampleScale]{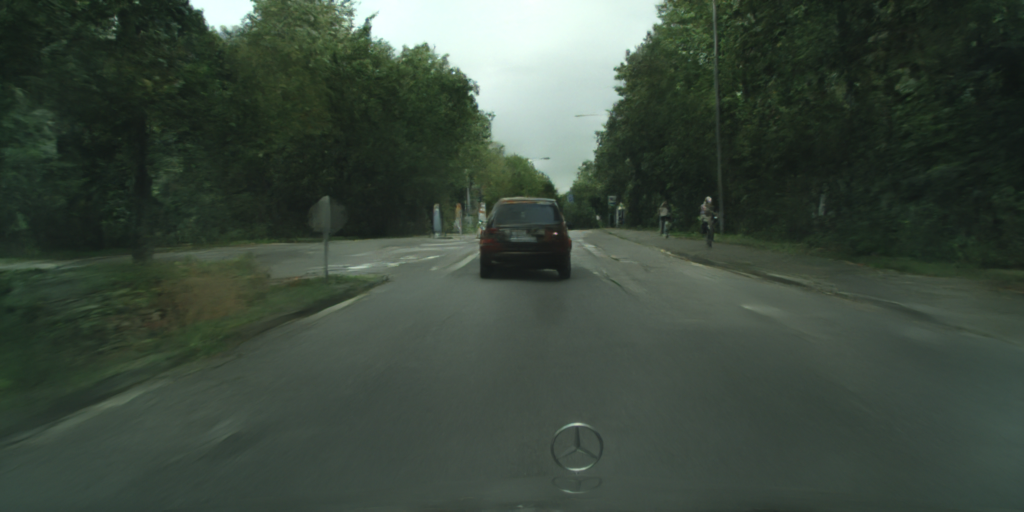}}
    \quad
    \subfloat{\includegraphics[scale=\GTLargeScale]{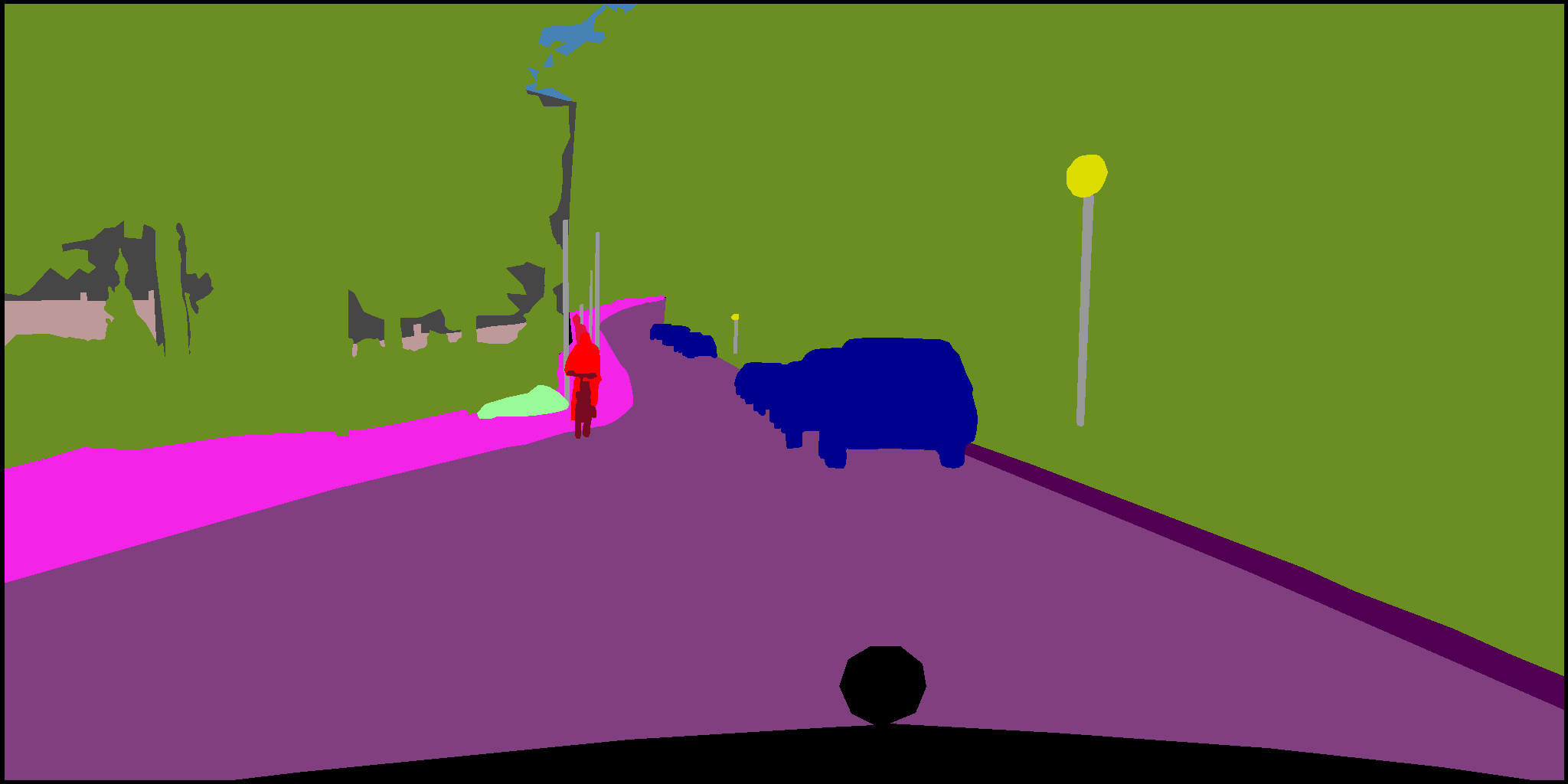}}
    \hfill
    \subfloat{\includegraphics[scale=\SingleSampleScale]{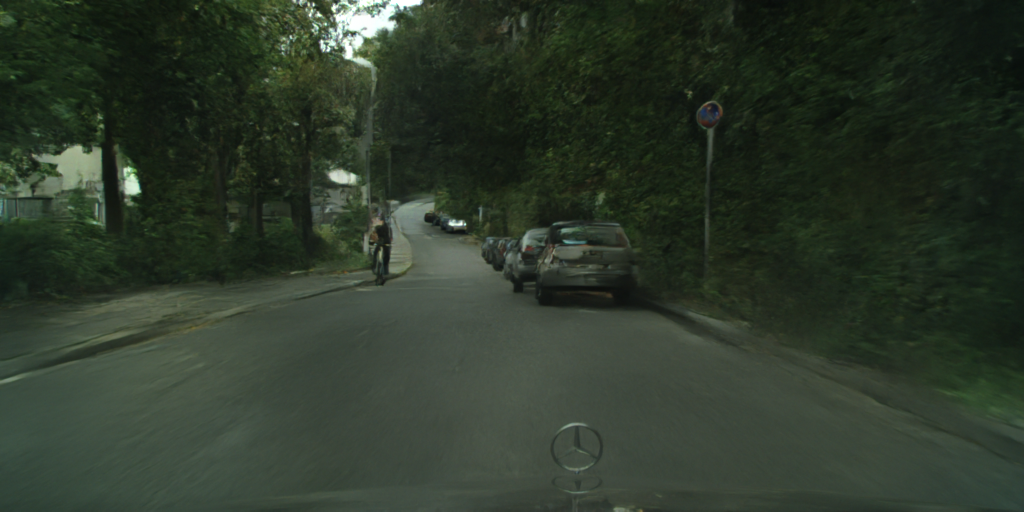}}
    \quad
    \subfloat{\includegraphics[scale=\GTLargeScale]{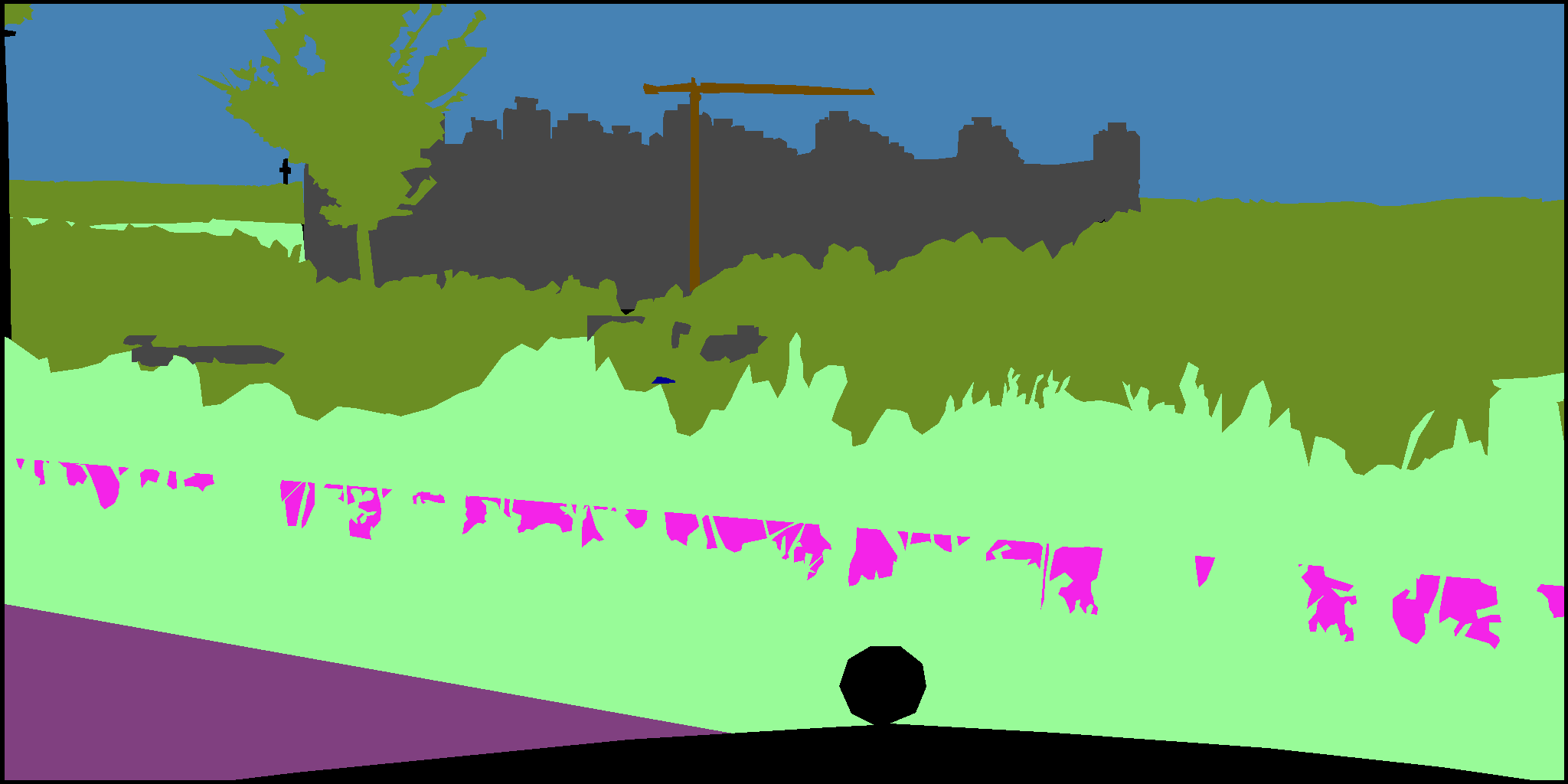}}
    \hfill
    \subfloat{\includegraphics[scale=\SingleSampleScale]{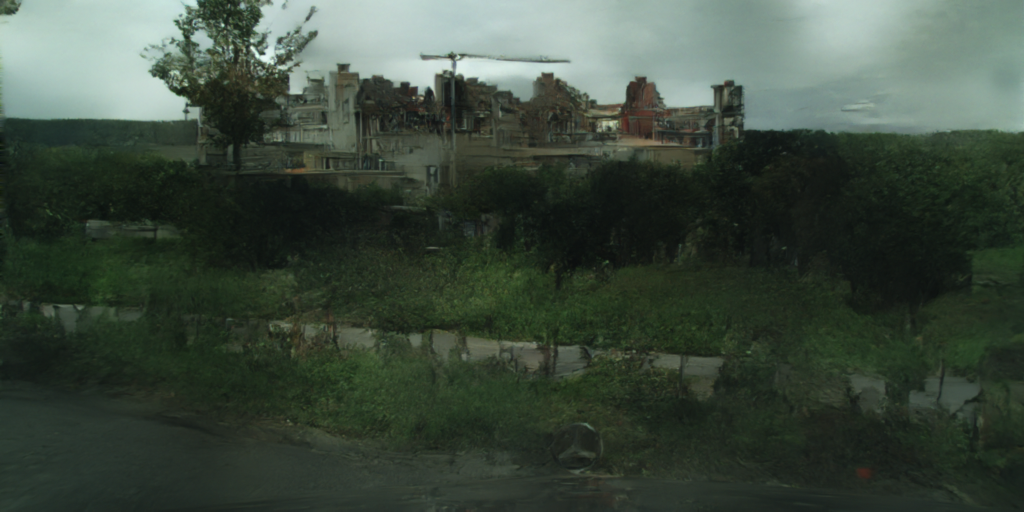}}
    \quad
}

\usepackage{graphicx}
\usepackage{comment}
\usepackage{amsmath,amssymb}
\usepackage{color}
\usepackage{url}
\usepackage{hyperref}

\newif\ifreview
\reviewfalse

\ifreview
	\usepackage{lineno}

	\linenumbers
\fi

\begin{document}


\def\SubNumber{084}

\def\GCPRTrack{Regular Track}

\title{Full-Glow: Fully conditional Glow for more realistic image generation}

\ifreview
\else

	\author{Moein Sorkhei\inst{1}\orcidID{0000-0001-6204-0778} \and
	Gustav Eje Henter\inst{1}\orcidID{0000-0002-1643-1054} \and
	Hedvig Kjellstr{\"o}m\inst{1,2}\orcidID{0000-0002-5750-9655}}
	
	\authorrunning{M.~Sorkhei et al.}
	
	\institute{KTH Royal Institute of Technology, Sweden
	\and Silo AI, Sweden\\
	\email{\{sorkhei,ghe,hedvig\}@kth.se} }
	
\fi

\maketitle              

\begin{abstract}
Autonomous agents, such as driverless cars, require large amounts of labeled visual data for their training. A viable approach for acquiring such data is training a generative model with collected real data, and then augmenting the collected real dataset with synthetic images from the model, generated with control of the scene layout and ground truth labeling. In this paper we propose {\em Full-Glow}, a fully conditional Glow-based architecture for generating plausible and realistic images of novel street scenes given a semantic segmentation map indicating the scene layout. Benchmark comparisons show our model to outperform recent works in terms of the semantic segmentation performance of a pretrained PSPNet. This indicates that images from our model are, to a higher degree than from other models, similar to real images of the same kinds of scenes and objects, making them suitable as training data for a visual semantic segmentation or object recognition system. 

\end{abstract}
%
%
%
\section{Introduction}
\label{sec:intro}
Autonomous mobile agents, such as driverless cars,
will be a cornerstone of the smart society of the future. Currently available datasets of labeled street scene images, such as Cityscapes \cite{Cordts2016Cityscapes}, are an important step in this direction, and could, e.g., be used for training models for semantic image segmentation. 
However, collecting such data poses challenges including privacy intrusions, the need for accurate crowd-sourced labels, and the requirement
to cover a huge state-space of different situations and environments. 
Another approach -- especially useful to gather data representing dangerous situations such as collisions with pedestrians -- is to generate training images with known ground-truth labeling using game engines or other virtual worlds, but this approach requires object and state-space variability to be manually engineered into the system.

A viable alternative to both these approaches 
is to augment existing datasets with synthetically-generated novel datapoints, produced by generative image models trained on the existing data.
This builds on recent applications of generative models for a variety of 
tasks such as image style transfer \cite{Isola_2017_CVPR} and modality transfer in medical imaging \cite{journals/corr/abs-1908-08074}. 

Among currently-available deep generative approaches, GANs \cite{goodfellow2014generative} are probably the most widely used in image generation, owing to their achievements in synthesizing realistic high-resolution output with novel and rich detail \cite{brock2019large,karras2019style}. Auto-regressive architectures \cite{oord2016pixel, van2016conditional} are usually computationally demanding (not parallelizable) 
and not feasible for generating higher-resolution images. Image samples generated by early variants of VAEs \cite{kingma2013auto,rezende2014stochastic} tended to suffer from blurriness \cite{zhao2017towards}, although the realism of VAE output has improved in recent years \cite{razavi2019generating,vahdat2020nvae}.

This article considers normalizing flows \cite{dinh2014nice, dinh2016density}, 
a different model class 
of growing
interest.
With recent improvements such as \emph{Glow} \cite{kingma2018glow}, flows can generate images with a quality that approaches that produced by GANs. Flows have also achieved competitive results in other tasks such as audio and video generation \cite{kim2018flowavenet, prenger2019waveglow, kumar2019videoflow}. Flow-based models exhibit 
several benefits 
compared to
GANs: 1) stable, monotonic training, 2) learning an explicit representation useful for down-stream tasks such as style transfer, 3) efficient synthesis, and 4) exact likelihood evaluation that could be used for density estimation.

In this paper, we propose a new, fully conditional Glow-based architecture called {\em Full-Glow} for generating plausible street scene images conditioned on the structure of the image content (i.e., the segmentation mask).
We show that, by using this model, we are able to synthesize moderately high-resolution images that are consistent with the given structure but differ substantially from the existing ground-truth images.
A quantitative comparison against previously proposed Glow-based models \cite{lu2020structured, journals/corr/abs-1908-08074} and the popular GAN-based conditional image-generation model pix2pix \cite{Isola_2017_CVPR}, finds that our improved conditioning allows us to synthesize images that achieve better semantic classification scores under a pre-trained semantic classifier.
We also provide visual comparisons of samples generated by different models. 

The remainder of this article is laid out as follows: Section \ref{sec:related} presents prior work in street-scene generation and image-to-image translation, while Section \ref{sec:flows} provides technical background on normalizing flows. Our proposed fully-conditional architecture is then introduced in Section \ref{sec:conditional} and validated experimentally in Section \ref{sec:exp}. 

\section{Related work}
\label{sec:related}

\paragraph{Synthetic data generation.} Street-scene image datasets such as Cityscapes \cite{Cordts2016Cityscapes}, CamVid \cite{brostow2009semantic}, and the KITTI dataset \cite{geiger2013vision} are useful for training 
vision systems for street-scene understanding. However, collecting and labeling such data is costly, resource demanding, and associated with privacy issues.
An effective alternative that allows for ground-truth labels and scene layout control 
is synthetic data generation using game engines 
\cite{tsirikoglou2017procedural, ros2016synthia, richter2016playing}. 
Despite these advantages, images generated by game engines tend to differ significantly from real-world images and may not always act as a replacement for real data.
Moreover, game engines generally only synthesize objects from pre-generated assets or recipes, meaning that variation has to be hand-engineered in. It is therefore difficult and costly to obtain diverse data in this manner.
Data generated from approaches such as ours address these shortcomings while maintaining the benefits of ground-truth labeling and scene layout control.

\vspace{-5mm}
\paragraph{Image-to-image translation.} 
In order to generate images for data-augmentation of supervised learning tasks, it is necessary to condition the image generation on an input, such that the ground-truth labeling of the generated image is known. For street-scene understanding, this conditioning takes the form of per-pixel class labels (a \emph{segmentation mask}), meaning that the augmentation task can be formulated as an image-to-image translation problem.
GANs \cite{goodfellow2014generative} have been employed for both paired and unpaired image-to-image translation problems \cite{Isola_2017_CVPR, Zhu_2017_ICCV}. While GANs can generate convincing-looking images, they are known to suffer from mode collapse and low output diversity \cite{grover2018flow}. Consequentially, their value in augmenting dataset diversity may be limited.

Likelihood-based models, on the other hand, explicitly aim to learn the probability distribution of the data. These models generally prefer sample diversity, sometimes at the expense of sample quality \cite{dinh2016density}, which has been linked to the mass-covering property of the likelihood objective \cite{minka2005divergence,theis2016note}.
Like for GANs \cite{brock2019large}, perceived image quality can often be improved by reducing the entropy of the distribution at synthesis time, relative to the distribution learned during training, cf.\ \cite{kingma2018glow,vahdat2020nvae}.
Flow-based models are a particular class of likelihood-based model that have gained recent attention after an architecture called Glow \cite{kingma2018glow} demonstrated impressive performance in unconditional image generation.
Previous works have applied flow-based models for image colorization \cite{ardizzone2019guided,ardizzone2020conditional}, image segmentation \cite{lu2020structured}, modality transfer in medical imaging \cite{journals/corr/abs-1908-08074}, and generating point clouds and image-to-image translation \cite{journals/corr/abs-1912-07009}.

So far, Glow-based models proposed for image-to-image translation \cite{lu2020structured, journals/corr/abs-1908-08074, journals/corr/abs-1912-07009} have only considered low-resolution tasks. Although the results are promising, they do not assess the full capacity of Glow for generating realistic image detail, for example in street scenes. High-resolution street-scene synthesis has been performed by the GAN-based model 
pix2pixHD \cite{wang2018high} on a GPU with very high memory capacity (24 GB).
In the present work, we synthesize moderately high resolution street scene images using a GPU with lower memory capacity (11--12 GB).
We extend previous works on Glow-based models by introducing a fully conditional architecture, and also by modeling high-resolution street-scene images, which is a more challenging task than the low-resolution output considered in prior work.

\section{Flow-based generative models}
\label{sec:flows}
Normalizing flows \cite{papamakarios2019normalizing} are a class of probabilistic generative models, able to represent complex probability densities in a manner that allows both easy sampling and efficient training based on explicit likelihood maximization.
The key idea is to use
a sequence of \emph{invertible} and \emph{differentiable} functions/transformations which (nonlinearly) transform a random variable $\mb{z}$ with a simple density function to another random variable $\mx$ with a more complex density function (and vice versa, thanks to invertibility):
\begin{equation}
    \mx \xleftrightarrow{\mb{f}_1} \mb{h}_1 \xleftrightarrow{\mb{f}_2} \mb{h}_2 
    \xleftrightarrow{\mb{f}_3}... 
    \xleftrightarrow{\mb{f}_K} \mb{z}
    \text{\gh{.}}
    \label{eq:flow}
\end{equation}
Each component transformation $\mb{f}_i$ is called a \emph{flow step}. The distribution of $\mb{z}$ (termed the \emph{latent}, \emph{source}, or \emph{base distribution}) is assumed to have a simple parametric form, such as an isotropic unit Gaussian.
Similar to in GANs, the generative process can be formulated as:
\vspace{-2mm}
\begin{eqnarray}
    \mz & \sim & p_{\mathrm{z}} (\mb{z}), \\
    \mx & = & \g_{\boldsymbol{\theta}}(\mz) = \f^{-1}_{\boldsymbol{\theta}}(\mz)
\end{eqnarray}
where $\mz$ is sampled from the base distribution and
\gh{$\g_{\boldsymbol{\theta}}$ represents the cumulative effect of the parametric invertible transformations in Equation \eqref{eq:flow}.}
The log-density function of $\mx$ under this transformation can be written as:
\begin{equation} \label{eq:change}
    \m{log} \, p_{\mathrm{x}} (\mx) = 
    \m{log} \, p_{\mathrm{z}} (\mb{z}) +
    \sum_{i=1}^{K} \m{log} \left|\m{det}\frac{\mathrm{d}\mb{h}_i}{\mathrm{d}\mb{h}_{i-1}}\right|
\end{equation}
\gh{using the \emph{change-of-variables theorem},}
where $\mb{h}_0 \triangleq \mx$ and $\mb{h}_K \triangleq \mb{z}$. Equation \eqref{eq:change} can be used to compute the exact dataset log-likelihood (not possible in GANs) and is 
the sole objective function for training flow-based models. 

The central design challenge of normalizing flows is to create expressive invertible transformations (typically parameterized by deep neural networks) where the so-called \emph{Jacobian log-determinant} in Equation \eqref{eq:change} remains computationally feasible to evaluate.
Often, this is achieved by designing transformations whose Jacobian matrix is triangular, making the determinant trivial to compute.
An important example
is NICE \cite{dinh2014nice}. NICE introduced the \emph{coupling layer}, which is a particular kind of flow nonlinearity that uses a neural network to invertibly transform half of the elements in $\mb{h}_k$ with respect to the other half. 
RealNVP \cite{dinh2016density} improved on this architecture using more general invertible transformations in the coupling layer and by imposing a hierarchical structure where the flow is partitioned into \emph{blocks} that operate at different resolutions. This hierarchy allows using smaller $\mb{z}$-vectors at the initial, smaller resolutions, speeding up computation, and 
has lately \hk{been used} by other prominent image-generation systems \cite{razavi2019generating,vahdat2020nvae}.
Glow \cite{kingma2018glow} added \emph{actnorm} as a replacement for batchnorm \cite{ioffe2015batch} and introduced invertible $1\times1$ convolutions to more efficiently mix variables in between the couplings.

A number of Glow-based architectures have 
been proposed for conditional image generation. In 
these models, the goal is to learn a distribution over the target image $\xb$ conditioned on the source image $\xa$. C-Glow \cite{lu2020structured} is based on the standard Glow architecture from \cite{kingma2018glow}, but makes all sub-steps inside the Glow conditional on the raw 
conditioning image $\xa$. The Dual-Glow \cite{journals/corr/abs-1908-08074} architecture instead builds a generative model of both source and target image together. It consists of two Glows where the base variables $\mb{z}_a$ of the source-image Glow determine the Gaussian distribution of the corresponding base variables $\mb{z}_b$ of the target-image Glow through a neural network. Because of the hierarchical structure of Glow, several different conditioning networks are used, one for each block of flow steps. C-Flow \cite{journals/corr/abs-1912-07009} described a similar structure of side-by-side Glows, but kept the Gaussian base distributions in the two flows independent. Instead, they used the latent variables $\mb{h}_{a,i}$ at every flow step $i$ of the \ms{target}-domain Glow to condition the transformation in the coupling layer at the corresponding level in the \ms{source}-domain Glow. Compared to the raw image-data conditioning in C-Glow, Dual-Glow and C-Flow simplify the conditional mapping task at the different levels since the source and target information sit at comparable levels of abstraction.

\section{Fully conditional Glow for scene generation}
\label{sec:conditional}
\begin{figure}[t]
    \centering
    \includegraphics[scale=0.15]{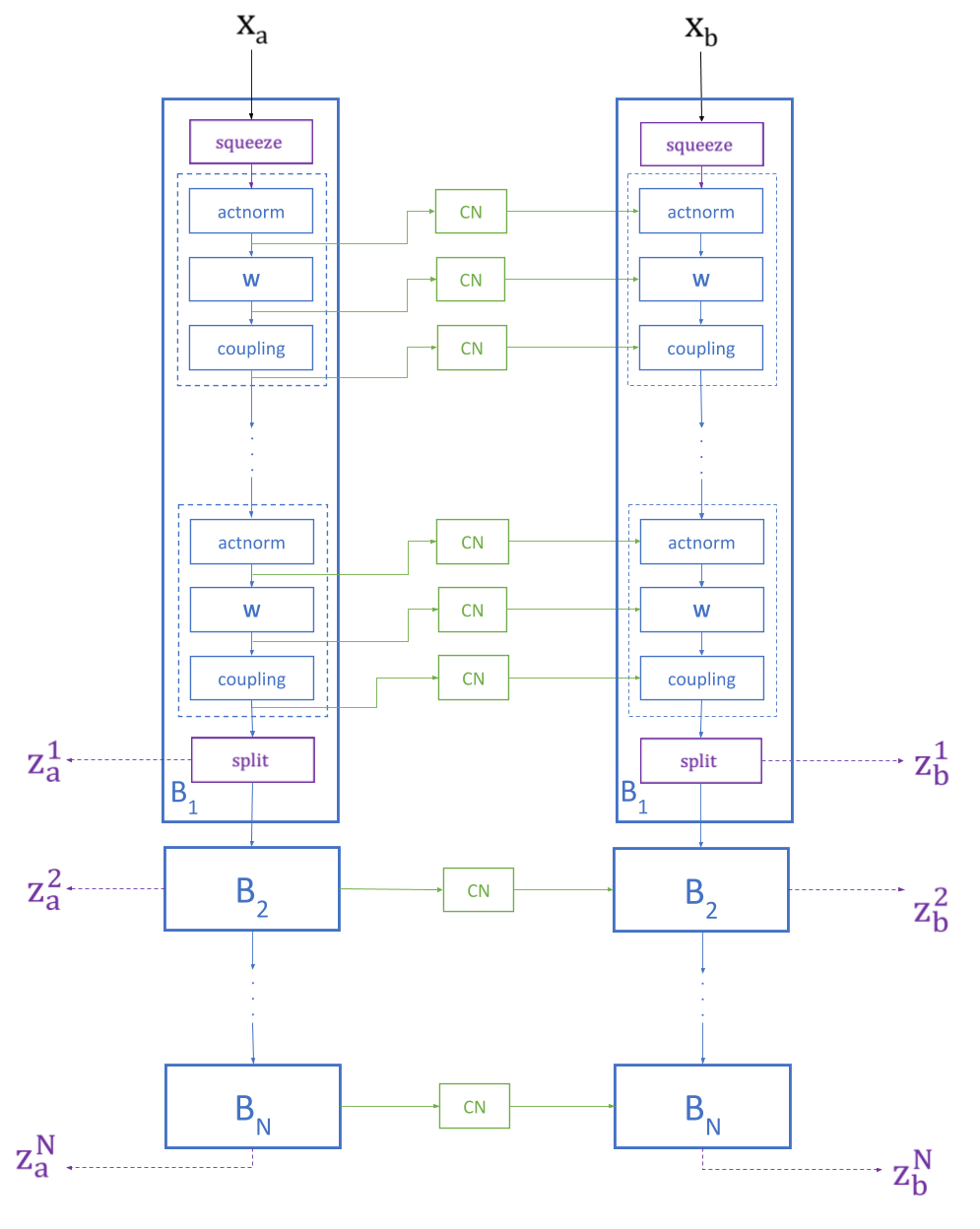}
    \caption{The proposed architecture, \gh{where all substeps have been made} conditional by inserting conditioning networks. $\mathbf{x}_{a}$ and $\mathbf{x}_{b}$ are pair\gh{ed images} in the source and target domains, respectively.}
    \label{fig:architecture}
\end{figure}

\gh{This section introduces our new, fully conditional Glow architecture for image-to-image translation, which combines key innovations from all three previous architectures, C-Glow, Dual-Glow, and C-Flow:}
\gh{Like Dual-Glow and C-Flow (but unlike C-Glow) we use two parallel stacks of Glow, so that we can leverage conditioning information at the relevant level of hierarchy and not be restricted to always use the raw source image as input. In contrast to Dual-Glow and C-Flow (but reminiscent of C-Glow), we introduce conditioning networks that allow \emph{all} operations in the target-domain Glow conditional on the source-domain information. The resulting architecture is illustrated in Figure \ref{fig:architecture}. Because of its fully conditional nature, we dub this architecture \emph{Full-Glow}.}


In \gh{our proposed} architecture, not only \gh{is} the coupling layer conditioned on the output of the corresponding operation in the source Glow, but \gh{the} actnorm and \gh{the} $1 \times 1$ convolutions in the target Glow are \gh{also}
connected to the source Glow. \gh{In particular}, the parameters of \gh{these two operations} in each \gh{target-side step} are generated by conditioning networks \gh{CN built from} convolutional layers followed by fully connected layers.
\gh{These} networks also \gh{enable} us to exploit other side information for conditioning\gh{, for instance} by concatenating the side information with the \gh{other} input features \gh{of} each conditioning network. We experimentally show that making the model fully conditional indeed allows for learning a better conditional distribution (measured with lower conditional bits-per-dimension) and more semantically meaningful images (measured using a pre-trained semantic classifier).

We will now describe the architecture of the fully-conditional target-domain Glow in more detail. We describe the computational statistical inference ($\f_{\boldsymbol{\theta}}$), but every transformation is invertible for synthesis ($\g_{\boldsymbol{\theta}}$) given the conditioning image $\xa$.

\vspace{-3mm}
\paragraph{Conditional actnorm.} The shift \gh{$\mathbf{t}$} and scale \gh{$\mathbf{s}$} parameters of \gh{the} conditional actnorm are computed as follows:
\vspace{-1mm}
\begin{equation}
    \mathbf{s}, \mathbf{t} = \texttt{CN}\left(\mathbf{x}_{\mathrm{act}}^{\mathrm{source}}\right)
\end{equation}
where $\mathbf{x}_{\mathrm{act}}^{\mathrm{source}}$ is the output of the corresponding actnorm in the source Glow.
For initializing the actnorm conditioning network (CN), we set all parameters of the network except those of \gh{the output} layer to \gh{small, random values}.
Similar to \cite{zhang2019fixup,kingma2018glow}, the weights of the \gh{output} layer are initialized to 0 and the biases are initialized \gh{such} that the output activations \ms{of the target side after applying actnorm} have mean 0 and std of 1 per channel for the first batch of data\gh{, similar to the scheme used for initialising actnorm in regular Glow}. 

\vspace{-3mm}
\paragraph{Conditional $1 \times 1$ convolution.} Like Glow, we represent the convolution kernel $W$ using an LU decomposition for easy log-determinant computation, but we have
conditioning networks generate the $\LL$, $\U$ matrices and the $\s$ vector:
\begin{equation}
    \LL, \U, \s = \texttt{CN}\left(\mathbf{x}_{\mathrm{\W}}^{\mathrm{source}}\right)
\end{equation}
where $\mathbf{x}_{\mathrm{\W}}^{\mathrm{source}}$ is the output of the corresponding 
\ms{$1\times1$ convolution} in the source Glow, \ms{$\LL$ is a lower triangular matrix with ones on the diagonal, $\U$ is an upper triangular matrix with zeros on the diagonal, and $\s$ is a vector}. Initialization again follows \cite{kingma2018glow}: we first sample a random rotation matrix $\W_0$ \gh{per layer, which we factorise using} the LU decomposition 
\ms{as $\W_0=\PP\LL_0\left(\U_0 + \mathrm{diag}(\s_0) \right)$}.
The conditioning network is then set up similar as for the actnorm, with weights and biases set so that its outputs on the first batch are constant and equivalent to the sampled rotation matrix $\W_0$.
The permutation matrix $\PP$ remains fixed throughout the optimization.

\vspace{-3mm} 
\paragraph{Conditional coupling layer.}
The conditional coupling layer resembles that of C-Flow (except that we use the whole source coupling output rather than half of it), where the network in the coupling layer takes input from both source and target sides:
\begin{eqnarray}
    \mathbf{x}_{1}^{\mathrm{target}}, \mathbf{x}_{2}^{\mathrm{target}} & = & \texttt{split}(\mathbf{x}^{\mathrm{target}}) \\
    \mathbf{o_1}, \mathbf{o_2} & = & \texttt{CN}\left(\mathbf{x}_{2}^{\mathrm{target}}, \mathbf{x}^{\mathrm{source}} \right) \\
    \ms{\mathbf{s}} & = & \ms{\mathrm{Sigmoid}(\mathbf{o_1 + 2})} \\
    \ms{\mathbf{t}} & = & \ms{\mathbf{o_2}}
\end{eqnarray}
where the \texttt{split} operation splits the input tensor along the channel dimension, $\mathbf{x}^{\mathrm{source}}$ is the output of the corresponding coupling layer in the source Glow, and $\mathbf{x}^{\mathrm{target}}$ is the output of the preceding $1 \times 1$ convolution in the target Glow. $\mathbf{s}$ and $\mathbf{t}$ are the affine coupling parameters.
The conditioning network inputs are concatenated channel-wise.

The objective function for the model has the same form as that of Dual-Glow:
\begin{equation}
    \label{eq:objective}
    \frac{1}{N} \left[ -\sum_{\gh{n}=1}^{N} \lambda \log p_{\boldsymbol{\theta}}\left(\mathbf{x}_{a}^{(\gh{n})}\right)-\sum_{\gh{n}=1}^{N} \log p_{\boldsymbol{\phi}}\left(\mathbf{x}_{b}^{(\gh{n})} \mid \mathbf{x}_{a}^{(\gh{n})}\right) \right]
\end{equation}
where $\boldsymbol{\theta}$ are the parameters of the source Glow and $\boldsymbol{\phi}$ are the parameters of the target Glow.
We note that there is one model (and term) for unconditional image generation in the source domain, coupled with a second model (and term) for conditional image generation in the target domain.
With the tuning parameter $\lambda$ set to unity, Equation \ref{eq:objective} is the joint likelihood of the source-target image pair $(\xa, \xb)$, and puts equal emphasis on learning to generate (and to normalize/analyze) both images.
In the limit $\lambda\to\infty$, we will learn an unconditional model of source images only. Using a
$\lambda$ below 1, however, helps the optimization process instead put more importance on the conditional distribution, which is our main priority in image-to-image translation. 
This ``exchange rate'' between bits of information in different domains is reminiscent of the tuning parameter in the information bottleneck principle \cite{tishby1999information}.

\section{Experiments}
\label{sec:exp}
This section reports our findings from applying the proposed model to the Cityscapes dataset from \cite{Cordts2016Cityscapes}.
Each data instance is a photo of a street scene that has has been segmented into objects of 30 different classes, such as road, sky, buildings, cars, and pedestrians. 5000 of these images come with \ms{fine} per-pixel class annotations of the image, a so called \emph{segmentation mask}. We used the data splits provided by the dataset (2975 training and 500 validation images), and trained a number of different models to generate street-scene images conditioned on their segmentation masks.

A common way to evaluate the quality of images generated based on the Cityscapes dataset is to apply well-known pre-trained classifiers such as FCN \cite{long2015fully} and (here) PSPNet \cite{zhao2017pyramid} to synthesized images (as done by \cite{Isola_2017_CVPR, wang2018high}). The idea is that if a synthesized image is of high quality, a classifier trained on real data should be able to successfully classify different objects in the synthetic image, and thus produce an estimated segmentation mask that closely agrees with the ground-truth segmentation mask. 
For likelihood-based models we also consider the conditional bits per dimension (BPD), $-\mathrm{log_2} \ p(\mathbf{x}_{b} | \mathbf{x}_{a})$, as a measure of
how well the conditional distribution learned by the model matches the real conditional distribution, when tested on held-out examples.

\vspace{-3mm}
\paragraph{Implementation details.}
\gh{Our main experiments were performed on images from the Cityscapes data down-sampled to $256 \times 256$ pixels \ms{(higher than C-Flow \cite{journals/corr/abs-1912-07009} that uses  64$\times$64 resolution)}.}
\gh{The Full-Glow model was implemented in PyTorch \cite{paszke2017automatic} and trained using the Adam optimizer \cite{2014adamOptimizer} with a learning rate of $10^{-4}$ and a batch size of 1.}
The conditioning networks (CN) for the actnorm and $1 \times 1$ convolution in our model consisted of three convolutional layers \gh{followed by} four fully connected layers. The CN for the coupling layer had two convolutional layers. \gh{Network weights} were initialized as described in Section \ref{sec:conditional}. We used $\lambda = 10^{-4}$ in the objective function Equation \ref{eq:objective}.
Training was consistently stable and monotonic; see the loss curve in the supplement.
Our implementation could be found at: \url{https://github.com/MoeinSorkhei/glow2}.

\DrawComparisonTable

\subsection{Quantitative comparison with other models}
\label{ssec:models}
We compare the performance of our model against C-Glow \cite{lu2020structured} and \hk{Dual-Glow} \cite{journals/corr/abs-1908-08074} (two previously proposed Glow-based models) and pix2pix \cite{Isola_2017_CVPR}
a widely used
GAN-based model for image-to-image translation. 

Since C-Glow was proposed to deal with low-resolution images, the authors exploited deep conditioning network in their model. We could not use equally deep conditioning networks in this task because the images we would like to generate are of higher resolution ($256 \times 256$). To enable valid comparisons, we trained two versions of the their model. In the first version, we allowed the conditioning networks to be deeper while the Glow itself is shallower (3 Blocks each with 8 Flows). In the second version, the Glow model is deeper (4 Blocks and each with 16 Flows) but the conditioning networks are shallower. More details about the models and their hyper-parameters can be found in the supplementary material. Note that the Glow models in C-Glow version 2, \hk{Dual-Glow}, and our model are all equally deep (4 Blocks and each having 16 Flows). All models, including Full-Glow, were trained for $\sim45$ epochs using the same training procedure described earlier.\footnote{We used these repositories \href{https://github.com/yolu1055/conditional-glow}{1}, \href{https://github.com/haolsun/dual-glow}{2}, \href{https://github.com/junyanz/pytorch-CycleGAN-and-pix2pix}{3} to obtain the official implementations of C-Glow, \hk{Dual-Glow}, and pix2pix. We did not find any official implementation for C-Flow.}

We \ms{sampled} from each trained model 3 times on the validation set, evaluated the synthesized images using PSPNet \cite{zhao2017pyramid}, \ms{and 
calculated the mean and standard deviation of the performance (denoted by $\pm$)}. \ms{The metrics used for evaluation are mean pixel accuracy, mean class accuracy, and mean intersection over union (IoU), as formulated in} \cite{long2015fully}. \ms{The mean pixel accuracy essentially computes mean accuracy over all the pixels of an image (which could easily be dominated by the sky, trees, and large objects that are mostly classified correctly.) Mean class accuracy, however, calculates the accuracy over the pixels of each class, and then takes average over different classes (where all classes are treated equally). Finally, mean class IoU calculates for each class the intersection over union for the objects of that class segmented in the synthesized image compared against the objects in the ground-truth segmentation. Optimally, this number should be 1, signifying complete overlap between segmented and ground-truth objects.}

Quantitative results of applying each model to the Cityscapes dataset in \gh{the} label $\rightarrow$ photo direction could be seen in Table \ref{tab:models_comparison}.  \ms{The results show that street scene images generated by Full-Glow are of higher quality from the viewpoint of semantic segmentation. The noticeable difference in classification performance confirms that the objects in the images generated by our model are more easily \textit{distinguishable} by the off-the-shelf semantic classifier. We attribute this to the fact that making the model fully conditional enables the target Glow to exploit the information available in the source image and to synthesize an image that follows the structure most.}


\begin{figure*}[!t]
    \DrawBlockTwo \quad
    \BlocksCaption
    \label{fig:models_compariso_all}
\end{figure*}

\subsection{Visual comparison with other models}
It is interesting to see how samples generated by different models are different visually. Figure \ref{fig:models_compariso_all} illustrates samples from different models given the same condition. An immediate observation is that C-Glow v.1 \cite{lu2020structured} (which has deeper conditioning networks but shallower Glow) is essentially unable to generate any meaningful image. \hk{Dual-Glow} \cite{journals/corr/abs-1908-08074}, however, is able to generate plausible images. Samples generated by pix2pix \cite{Isola_2017_CVPR} exhibit vibrant colors (especially for the buildings) but the important objects (such as cars) that constitute the general structure of the image are sometimes distorted. We believe this is the reason behind scoring low with the semantic classifier. Respecting the structure seems to be more important than having vibrant colors in order to get higher classification accuracy. Different samples taken from our model show the benefit of flow-based models in synthesizing different images every time we sample. Most of the difference can be seen in the colors of the objects such as cars.

Generally, the samples generated by likelihood-based models appear dimmed to some extent. This is in contrast with GAN-based samples, which often have realistic colors. This is probably related to the fundamental difference in the optimization process of the two categories. GAN-based models tend to collapse to regions of data where only plausible samples could have come from, and they might not have support over other data regions \cite{grover2018flow} -- also seen as lack of diversity in their samples. In contrast, likelihood-based models try to learn a distribution that has support over \ms{wider} data regions while maximizing the probability of the available datapoints. The latter approach seems to result in generating samples that are diverse but have somewhat muted colors (especially with lower temperatures).



\DrawTempTable

\subsection{Effect of temperature}
\gh{As noted above,} likelihood-based models such as Glow \cite{kingma2018glow} generally tend to \emph{overestimate} the \gh{variability of the} data distribution \cite{theis2016note,minka2005divergence}, hence occasionally generating \gh{implausible output} samples. A common way to circumvent this issue is to reduce the \gh{diversity of the output at generation time. For flows, this can be done by reducing the standard deviation} of the base distribution \gh{by a factor $T$} (known as \gh{the} temperature)\gh{. While $T=1$ corresponds to sampling from the estimated maximum-likelihood distribution, reducing $T$ generally} results in the \gh{output} distribution
\gh{becoming} concentrated \gh{on a core region of especially-probable output samples. Similar ideas are widely used not only in flow-based models (cf.\ \cite{kingma2018glow}) but also in other generative models such as GANs, VAEs, and Transformer-based language models \cite{brock2019large,vahdat2020nvae,holtzman2020curious,brown2020language}.}

We investigated the effect of temperature by evaluating the performance of the model \gh{on} samples generated \gh{at} different temperatures \gh{(instead of $T=1$ as in previous experiments)}. We \ms{sampled} on the validation set 3 times with the trained model and evaluated 
using the PSPNet semantic classifier \gh{as before}. \gh{The results are reported in} Table \ref{tab:temp_effect} \gh{and} suggest that the optimal temperature is around 0.8 \gh{for} this task.
\gh{That setting strikes a compromise where} colors are vibrant \gh{while object} structure \gh{is} well-maintained\gh{, enabling} the classifier \gh{to well} understand the objects in the synthesized image. \gh{Also note} the small \gh{standard deviation at} lower temperatures, which \gh{agrees with our expectation that inter-sample variability would be small at low temperature. Example images generated at different temperatures are provided}
in the supplementary material.

\begin{figure*}[!t]
    \centering
    \ContentTransferText
    \DrawContentTransferOne
    \DrawContentTransferTwo
    \DrawContentTransferThree
    \ContentTransferCaption
    \label{fig:new_content_all}
\end{figure*}

\subsection{Content transfer}
Style transfer in general is an interesting application for synthesizing new images, which incorporates transferring the style of an image into a new structure (condition). In this experiment, we try to transfer the \textit{content} of a real photo to a new structure (segmentation). Previous work \cite{ardizzone2020conditional, journals/corr/abs-1912-07009} \gh{has performed} similar experiments with flow-based models \gh{but either} on a different dataset or in very low resolution (\ms{64$\times$64}).
We \gh{demonstrate} how the learned representation enables us to synthesize an image given a desired content and a desired structure in relatively high resolution while maintaining the details of the content. 

Suppose $\xB{1}$ is the image with the desired content (with $\xA{1}$ being its segmentation) and $\xA{2}$ is the new segmentation to which we are interested to transfer the content. We can take the following steps ($\mathbf{g}_{\boldsymbol{\theta}}(.)$ and $\mathbf{f}_{\phi}(.)$ are the forward functions in the source and target Glows respectively):
\begin{enumerate}
    \item Extract representation of the desired content: $\mathbf{z}_{\mathrm{b}}^{1}=\mathbf{f}_{\phi}\left(\mathbf{x}_{\mathrm{b}}^{1} \: | \: \mathbf{g}_{\boldsymbol{\theta}}\left(\mathbf{x}_{\mathrm{a}}^{1}\right) \right)$.
    
    \item Apply the content to the new segmentations: $\mathbf{x}_{\mathrm{b}}^{\mathrm{new}}=\mathbf{f}_{\phi}^{-1} \left(\mathbf{z}_{\mathrm{b}}^{1} \: | \: \mathbf{g}_{\boldsymbol{\theta}}\left(\mathbf{x}_{\mathrm{a}}^{2}\right) \right)$. 
\end{enumerate}

Figure \ref{fig:new_content_all} shows examples of transferring the content of an image to another segmentation. We can see that the model is able to successfully apply the content of large objects such as buildings, trees, cars to the \gh{desired image} structure. 
\ms{So often a given content and a given structure may not agree with each other.} For instance, when there are cars in the content which are missing in the segmentation or vice versa. This kind of \gh{mismatch} is quite common for \gh{content transfer on} Cityscapes images as these images have a lot of objects placed in different positions.
In such a case, the model tries to respect the structure as much as possible while filling it with the given content. \ms{The results show that content transfer is so useful in data augmentation since, given the desired content, the model can fill the structure with coherent information which makes the output image much realistic. This technique could practically be applied to any content and any structure (provided that they do not mismatch completely), hence enabling one to synthesize many more images.}


\begin{figure}[t]
    \centering
    \DrawDiverseOne
    \caption{\DiverseCaptionMultiple}
    \label{fig:diverse_1}
\end{figure}

\begin{figure}[t]
    \centering
    \DrawDiverseTwo
    \caption{\DiverseCaptionMultiple}
    \label{fig:diverse_2}
\end{figure}

\subsection{Higher-resolution samples}
In order to see how expressive the model is \gh{at even} higher resolutions, we trained the proposed model on 512$\times$1024 images.
Example output images from the trained model are provided in Figures \ref{fig:diverse_1} and 
\ref{fig:diverse_2}.
It is known that higher temperatures show more diversity but the structures become somewhat distorted. We chose to sample with temperature 0.9 for \gh{these} higher-resolution images.
The diversity \gh{between} multiple samples is especially obvious looking at the cars.
\gh{Additional} high-resolution samples \gh{are available} in the supplementary material.

\section{Conclusions}
\label{sec:conclusion}
In this paper, we proposed a fully conditional Glow-based architecture for more realistic conditional street-scene image generation. We quantitatively compared our method against previous work and observed that our improved conditioning allows for generating images that are more interpretable by the semantic classifier. We also used the architecture to synthesize higher-resolution images in order to better show diversity of samples and the capabilities of Glow at higher resolutions. In addition, we demonstrated how new meaningful images could be synthesized based on a desired content and a desired structure, which is a compelling option for high-quality data augmentation.

\vspace{-3mm}
\paragraph{Future work.} While our results are promising, further work remains to be done in order to close the gap between synthetic images and real-world photographs, for instance by adding self-attention \cite{ho2019flow++} and by leveraging approaches that combine the strong points of Full-Glow with the advantages of GANs, e.g., following \cite{grover2018flow,lucas2019adaptive}. That said, we believe the quality of the synthetic images is already at a level where it also is worth exploring their utility in training systems for autonomous cars and other mobile agents, which remains to be observed in future works.

\paragraph{Acknowledgments.} This work was partially funded by the KTH Digital Futures program and by the Wallenberg AI, Autonomous Systems and Software Program (WASP) funded by the Knut and Alice Wallenberg Foundation.
%
%
%
\bibliographystyle{splncs04}
\bibliography{egbib}

\begin{thebibliography}{10}
\providecommand{\url}[1]{\texttt{#1}}
\providecommand{\urlprefix}{URL }
\providecommand{\doi}[1]{https://doi.org/#1}

\bibitem{ardizzone2020conditional}
Ardizzone, L., Kruse, J., L{\"u}th, C., Bracher, N., Rother, C., K{\"o}the, U.:
  Conditional invertible neural networks for diverse image-to-image
  translation. In: DAGM German Conference on Pattern Recognition. pp. 373--387
  (2020)

\bibitem{ardizzone2019guided}
Ardizzone, L., L{\"u}th, C., Kruse, J., Rother, C., K{\"o}the, U.: Guided image
  generation with conditional invertible neural networks. arXiv preprint
  arXiv:1907.02392  (2019)

\bibitem{brock2019large}
Brock, A., Donahue, J., Simonyan, K.: Large scale {GAN} training for high
  fidelity natural image synthesis. In: International Conference on Learning
  Representations (2019)

\bibitem{brostow2009semantic}
Brostow, G.J., Fauqueur, J., Cipolla, R.: Semantic object classes in video: A
  high-definition ground truth database. Pattern Recognition Letters
  \textbf{30}(2),  88--97 (2009)

\bibitem{brown2020language}
Brown, T.B., Mann, B., Ryder, N., Subbiah, M., Kaplan, J.D., Dhariwal, P.,
  Neelakantan, A., Shyam, P., Sastry, G., Askell, A., Agarwal, S.,
  Herbert-Voss, A., Krueger, G., Henighan, T., Child, R., Ramesh, A., Ziegler,
  D., Wu, J., Winter, C., Hesse, C., Chen, M., Sigler, E., Litwin, M., Gray,
  S., Chess, B., Clark, J., Berner, C., McCandlish, S., Radford, A., Sutskever,
  I., Amodei, D.: Language models are few-shot learners. In: Neural Information
  Processing Systems. pp. 1877--1901 (2020)

\bibitem{Cordts2016Cityscapes}
Cordts, M., Omran, M., Ramos, S., Rehfeld, T., Enzweiler, M., Benenson, R.,
  Franke, U., Roth, S., Schiele, B.: The {C}ityscapes dataset for semantic
  urban scene understanding. In: IEEE Conference on Computer Vision and Pattern
  Recognition (2016)

\bibitem{dinh2014nice}
Dinh, L., Krueger, D., Bengio, Y.: {NICE}: Non-linear independent components
  estimation. arXiv preprint arXiv:1410.8516  (2014)

\bibitem{dinh2016density}
Dinh, L., Sohl-Dickstein, J., Bengio, S.: Density estimation using {Real NVP}.
  In: International Conference on Learning Representations (2017)

\bibitem{geiger2013vision}
Geiger, A., Lenz, P., Stiller, C., Urtasun, R.: Vision meets robotics: The
  {KITTI} dataset. International Journal of Robotics Research  \textbf{32}(11),
   1231--1237 (2013)

\bibitem{goodfellow2014generative}
Goodfellow, I., Pouget-Abadie, J., Mirza, M., Xu, B., Warde-Farley, D., Ozair,
  S., Courville, A., Bengio, Y.: Generative adversarial nets. In: Neural
  Information Processing Systems (2014)

\bibitem{grover2018flow}
Grover, A., Dhar, M., Ermon, S.: Flow-{GAN}: Combining maximum likelihood and
  adversarial learning in generative models. In: AAAI Conference on Artificial
  Intelligence (2018)

\bibitem{ho2019flow++}
Ho, J., Chen, X., Srinivas, A., Duan, Y., Abbeel, P.: Flow++: Improving
  flow-based generative models with variational dequantization and architecture
  design. arXiv preprint arXiv:1902.00275  (2019)

\bibitem{holtzman2020curious}
Holtzman, A., Buys, J., Du, L., Forbes, M., Choi, Y.: The curious case of
  neural text degeneration. In: International Conference on Learning
  Representations (2020)

\bibitem{huszar2017gaussian}
Husz\'{a}r, F.: Gaussian distributions are soap bubbles (2017),
  \href{https://www.inference.vc/high-dimensional-gaussian-distributions-are-soap-bubble/}{https://www.inference.vc/high-dimensional-gaussian-distributions-are-soap-bubbles}

\bibitem{ioffe2015batch}
Ioffe, S., Szegedy, C.: Batch normalization: Accelerating deep network training
  by reducing internal covariate shift. arXiv preprint arXiv:1502.03167  (2015)

\bibitem{Isola_2017_CVPR}
Isola, P., Zhu, J.Y., Zhou, T., Efros, A.A.: Image-to-image translation with
  conditional adversarial networks. In: IEEE Conference on Computer Vision and
  Pattern Recognition (2017)

\bibitem{karras2019style}
Karras, T., Laine, S., Aila, T.: A style-based generator architecture for
  generative adversarial networks. In: IEEE Conference on Computer Vision and
  Pattern Recognition (2019)

\bibitem{kim2018flowavenet}
Kim, S., Lee, S.g., Song, J., Kim, J., Yoon, S.: {FloWaveNet}: A generative
  flow for raw audio. In: International Conference on Machine Learning (2019)

\bibitem{2014adamOptimizer}
Kingma, D.P., Ba, J.: Adam: A method for stochastic optimization. arXiv
  preprint arXiv:1412.6980  (2014)

\bibitem{kingma2018glow}
Kingma, D.P., Dhariwal, P.: Glow: Generative flow with invertible 1x1
  convolutions. In: Neural Information Processing Systems (2018)

\bibitem{kingma2013auto}
Kingma, D.P., Welling, M.: Auto-encoding variational {Bayes}. In: International
  Conference on Learning Representations (2014)

\bibitem{kumar2019videoflow}
Kumar, M., Babaeizadeh, M., Erhan, D., Finn, C., Levine, S., Dinh, L., Kingma,
  D.: {VideoFlow}: A flow-based generative model for video. arXiv preprint
  arXiv:1903.01434  (2019)

\bibitem{long2015fully}
Long, J., Shelhamer, E., Darrell, T.: Fully convolutional networks for semantic
  segmentation. In: IEEE Conference on Computer Vision and Pattern Recognition
  (2015)

\bibitem{lu2020structured}
Lu, Y., Huang, B.: Structured output learning with conditional generative
  flows. In: AAAI Conference on Artificial Intelligence (2020)

\bibitem{lucas2019adaptive}
Lucas, T., Shmelkov, K., Alahari, K., Schmid, C., Verbeek, J.: Adaptive density
  estimation for generative models. In: Neural Information Processing Systems.
  pp. 11993--12003 (2019)

\bibitem{mescheder2018training}
Mescheder, L., Geiger, A., Nowozin, S.: Which training methods for {GAN}s do
  actually converge? In: International Conference on Machine Learning. pp.
  3481--3490 (2018)

\bibitem{minka2005divergence}
Minka, T.: Divergence measures and message passing. Tech. Rep. MSR-TR-2005-173,
  Microsoft Research, Cambridge, UK (2005)

\bibitem{van2016conditional}
van~den Oord, A., Kalchbrenner, N., Espeholt, L., Vinyals, O., Graves, A.,
  Kavukcuoglu, K.: Conditional image generation with {pixelCNN} decoders. In:
  Neural Information Processing Systems (2016)

\bibitem{vandenoord2018parallel}
van~den Oord, A., Li, Y., Babuschkin, I., Simonyan, K., et~al.: Parallel
  {W}ave{N}et: Fast high-fidelity speech synthesis. In: International
  Conference on Machine Learning. pp. 3918--3926 (2018)

\bibitem{oord2016pixel}
Oord, A.v.d., Kalchbrenner, N., Kavukcuoglu, K.: Pixel recurrent neural
  networks. arXiv preprint arXiv:1601.06759  (2016)

\bibitem{papamakarios2019normalizing}
Papamakarios, G., Nalisnick, E., Rezende, D.J., Mohamed, S., Lakshminarayanan,
  B.: Normalizing flows for probabilistic modeling and inference. arXiv
  preprint arXiv:1912.02762  (2019)

\bibitem{paszke2017automatic}
Paszke, A., Gross, S., Chintala, S., Chanan, G., Yang, E., DeVito, Z., Lin, Z.,
  Desmaison, A., Antiga, L., Lerer, A.: Automatic differentiation in {PyTorch}.
  In: NIPS 2017 Workshop Autodiff (2017)

\bibitem{prenger2019waveglow}
Prenger, R., Valle, R., Catanzaro, B.: {W}ave{G}low: A flow-based generative
  network for speech synthesis. In: IEEE International Conference on Acoustics,
  Speech and Signal Processing (2019)

\bibitem{journals/corr/abs-1912-07009}
Pumarola, A., Popov, S., Moreno-Noguer, F., Ferrari, V.: {C-Flow}: Conditional
  generative flow models for images and {3D} point clouds. In: IEEE Conference
  on Computer Vision and Pattern Recognition (2020)

\bibitem{razavi2019generating}
Razavi, A., van~den Oord, A., Vinyals, O.: Generating diverse high-fidelity
  images with {VQ-VAE-2}. In: Neural Information Processing Systems (2019)

\bibitem{rezende2014stochastic}
Rezende, D.J., Mohamed, S., Wierstra, D.: Stochastic backpropagation and
  approximate inference in deep generative models. In: International Conference
  on Machine Learning (2014)

\bibitem{richter2016playing}
Richter, S.R., Vineet, V., Roth, S., Koltun, V.: Playing for data: Ground truth
  from computer games. In: European Conference on Computer Vision (2016)

\bibitem{ros2016synthia}
Ros, G., Sellart, L., Materzynska, J., Vazquez, D., Lopez, A.M.: The {SYNTHIA}
  dataset: A large collection of synthetic images for semantic segmentation of
  urban scenes. In: IEEE Conference on Computer Vision and Pattern Recognition
  (2016)

\bibitem{checkpointing}
Salimans, T., Bulatov, Y.: Gradient checkpointing (2018),
  \href{https://github.com/openai/gradient-checkpointing}{https://github.com/openai/gradient-checkpointing}

\bibitem{journals/corr/abs-1908-08074}
Sun, H., Mehta, R., Zhou, H.H., Huang, Z., Johnson, S.C., Prabhakaran, V.,
  Singh, V.: {DUAL-GLOW}: Conditional flow-based generative model for modality
  transfer. In: IEEE International Conference on Computer Vision (2019)

\bibitem{theis2016note}
Theis, L., van~den Oord, A., Bethge, M.: A note on the evaluation of generative
  models. In: International Conference on Learning Representations (2016)

\bibitem{tishby1999information}
Tishby, N., Pereira, F.C., Bialek, W.: The information bottleneck method. In:
  Proceedings of the Allerton Conference on Communication, Control and
  Computing. vol.~37, pp. 368--377 (2000)

\bibitem{tsirikoglou2017procedural}
Tsirikoglou, A., Kronander, J., Wrenninge, M., Unger, J.: Procedural modeling
  and physically based rendering for synthetic data generation in automotive
  applications. arXiv preprint arXiv:1710.06270  (2017)

\bibitem{vahdat2020nvae}
Vahdat, A., Kautz, J.: {NVAE}: A deep hierarchical variational autoencoder.
  arXiv preprint arXiv:2007.03898  (2020)

\bibitem{wang2018high}
Wang, T.C., Liu, M.Y., Zhu, J.Y., Tao, A., Kautz, J., Catanzaro, B.:
  High-resolution image synthesis and semantic manipulation with conditional
  {GANs}. In: IEEE Conference on Computer Vision and Pattern Recognition (2018)

\bibitem{zhang2019fixup}
Zhang, H., Dauphin, Y.N., Ma, T.: Fixup initialization: {R}esidual learning
  without normalization. In: International Conference on Learning
  Representations (2019)

\bibitem{zhao2017pyramid}
Zhao, H., Shi, J., Qi, X., Wang, X., Jia, J.: Pyramid scene parsing network.
  In: IEEE Conference on Computer Vision and Pattern Recognition (2017)

\bibitem{zhao2017towards}
Zhao, S., Song, J., Ermon, S.: Towards deeper understanding of variational
  autoencoding models. arXiv preprint arXiv:1702.08658  (2017)

\bibitem{Zhu_2017_ICCV}
Zhu, J.Y., Park, T., Isola, P., Efros, A.A.: Unpaired image-to-image
  translation using cycle-consistent adversarial networks. In: IEEE
  International Conference on Computer Vision (2017)

\end{thebibliography}



\clearpage
\section*{Appendix}
\appendix

In this supplementary material we provide: \begin{itemize}
    \item additional examples of samples generated by different models and high-resolution examples (see Section \ref{additional} with Figures \ref{fig:models_compariso_all_OLD}, \ref{fig:single_1}, and \ref{fig:single_2}), 
    \item experiments with additional conditioning information and with deeper models (see Section \ref{deeper} with Figures \ref{fig:bmap} and \ref{fig:extended_arch} as well as Table \ref{tab:b_map_with_deep}), 
    \item visual examples of the effect of varying the temperature (see Section \ref{temperature} with Figures \ref{fig:temp_effect_1} and \ref{fig:temp_effect_2}),
    \item further implementation details of the models (see Section \ref{sec:models_details}),
    \item and finally an example loss curve from training our proposed model (see Section \ref{sec:loss_curve} with Figure \ref{fig:loss}).
\end{itemize}

\section{Additional samples}
\label{additional}
In the experiments in the paper, we compared the performance of the models quantitatively and showed a visual example of the samples generated by different models. Figure \ref{fig:models_compariso_all_OLD} here shows another visual example of samples generated by different models. 

We also showed that Full-Glow is able to generate plausible high-resolution images given a segmentation map as condition. Figures \ref{fig:single_1} and \ref{fig:single_2} show additional synthesized images in resolution 512$\times$1024 along with the conditions.

\begin{figure*}[]
    \DrawBlockThree
    \BlocksCaption
    \label{fig:models_compariso_all_OLD}
\end{figure*}

\begin{figure*}[!t]
    \centering
    \DrwaSingleOne
    \caption{\SingleCaption}
    \label{fig:single_1}
\end{figure*}

\begin{figure*}[t]
    \centering
    \DrwaSingleTwo
    \caption{\SingleCaption}
    \label{fig:single_2}
\end{figure*}

\section{Effect of boundary maps and deeper models}
\label{deeper}
This part illustrates the effect of using boundary maps and model depth on the performance of the model.

\begin{figure}[]
    \centering
    \includegraphics[scale=.08]{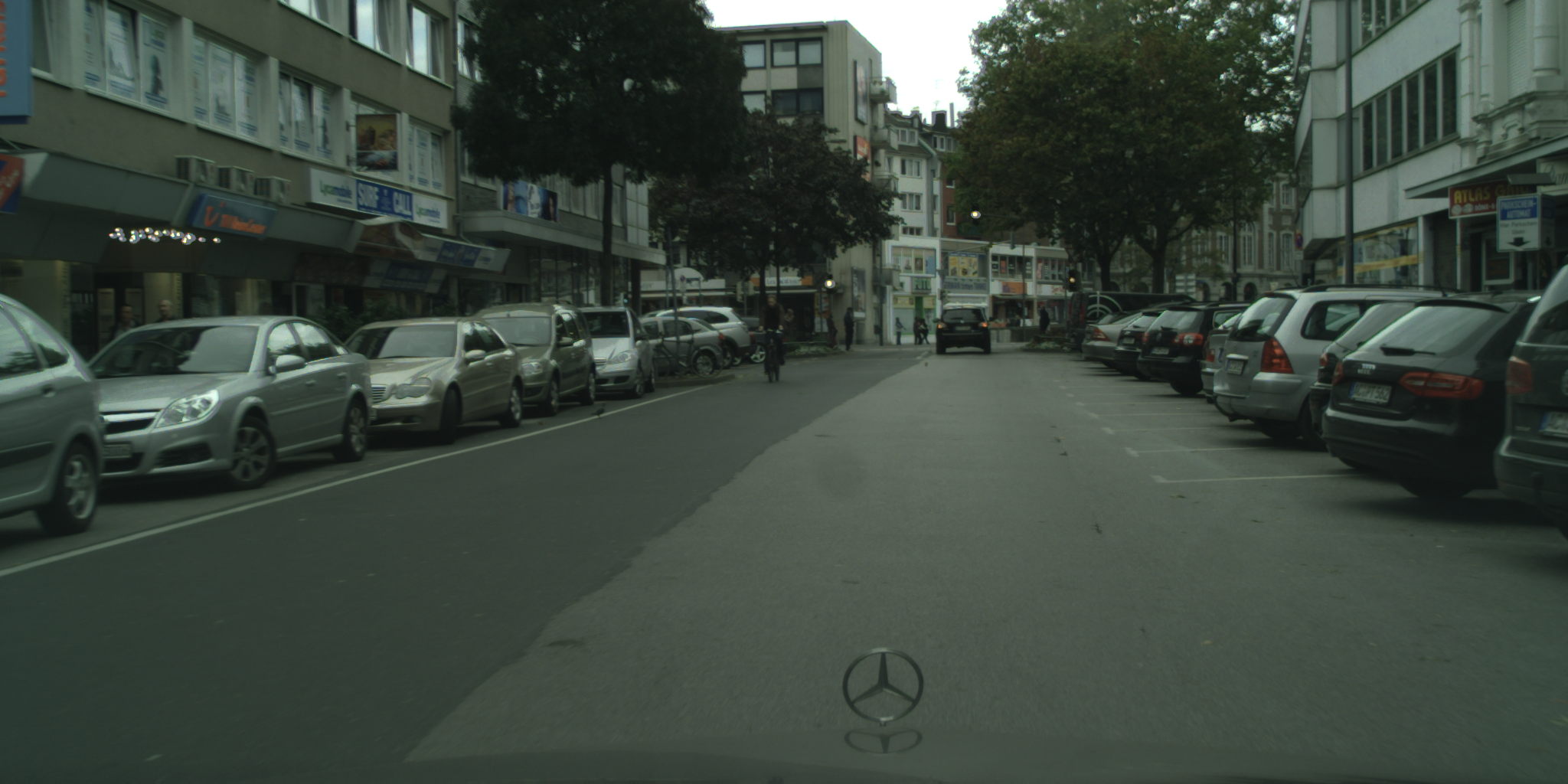}
    \includegraphics[scale=.08]{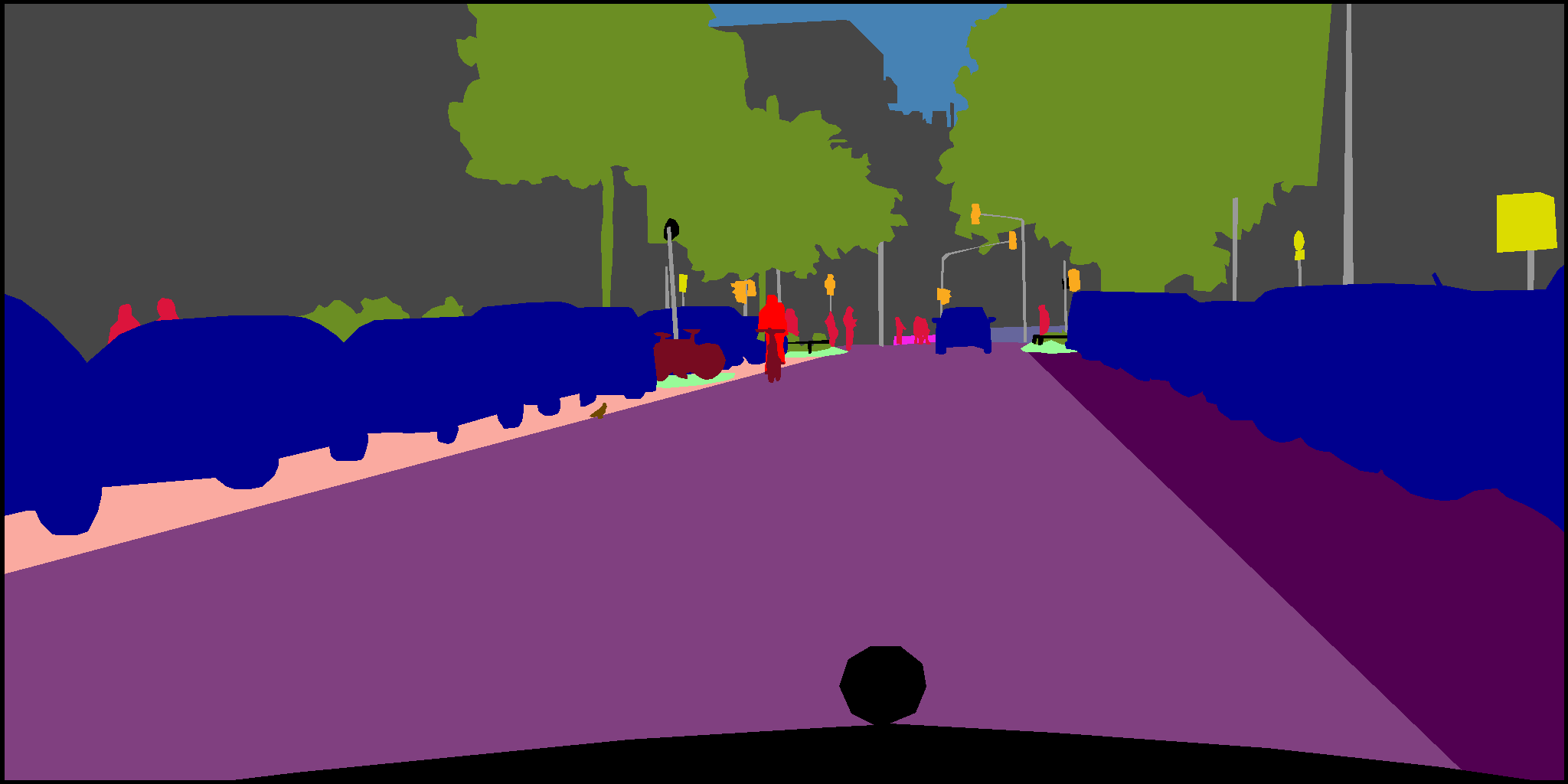}
    \includegraphics[scale=.08]{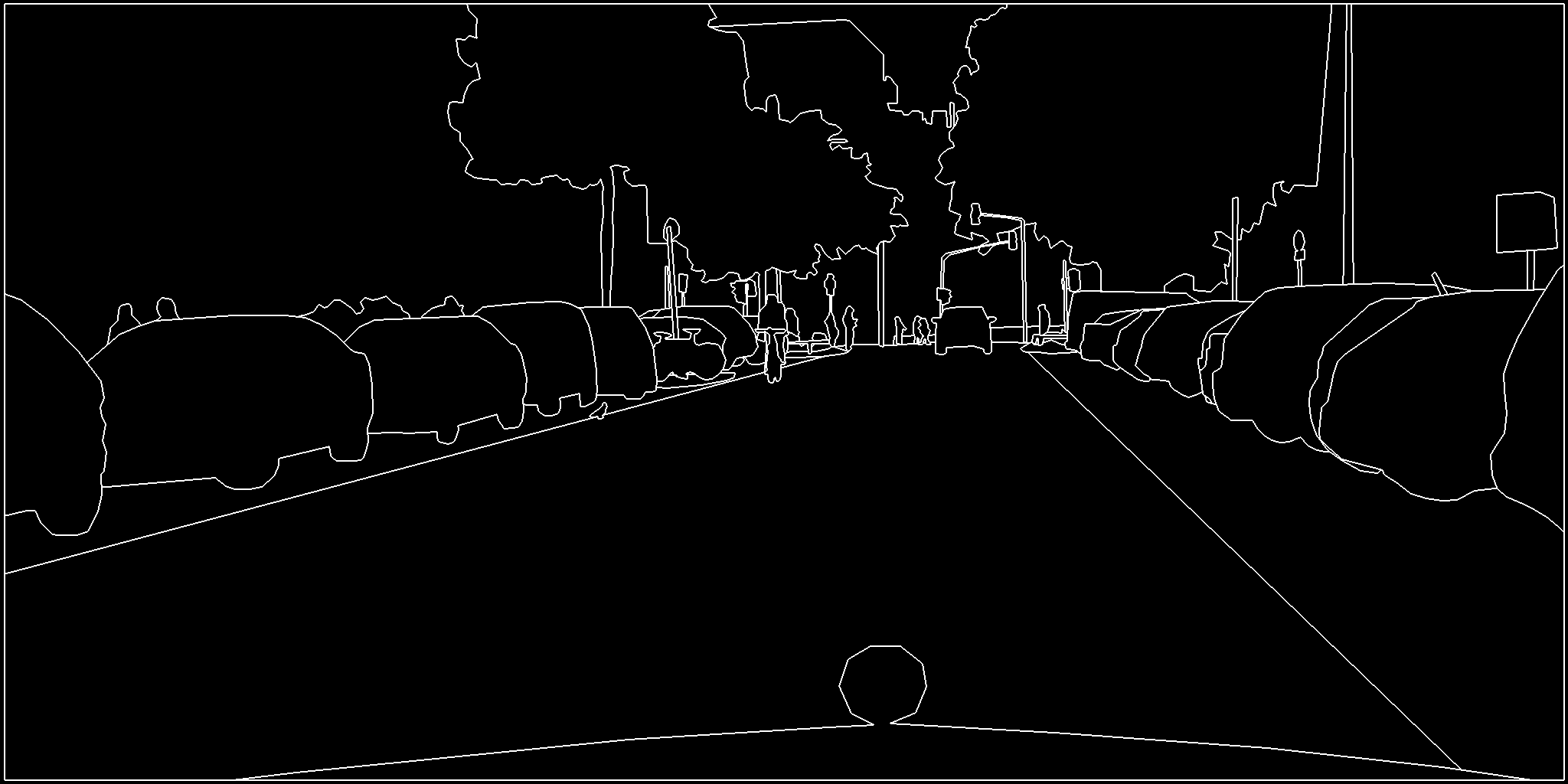}
    \caption{Different modalities of an image in Cityscapes \cite{Cordts2016Cityscapes}. Boundary maps are not explicitly available but could be derived from available instance IDs.}
    \label{fig:bmap}
\end{figure}

\vspace{-3mm}
\paragraph{Boundary maps.} 
When using segmentations as condition, the boundaries between objects of the same class are not visible when the objects are placed next to each other. This is because they all belong to the same class in the segmentation mask. As observed by \cite{wang2018high}, in the Cityscapes \cite{Cordts2016Cityscapes} dataset, each object in an image has an instance ID, which could be used to derive the boundary maps between the objects. For each pixel, if any of the 4 neighbors around it corresponds to a different instance ID than that of the pixel, the pixel will take value 1, otherwise it will be 0 (see Figure \ref{fig:bmap}).

\begin{figure}[]
    \centering
    \includegraphics[scale=0.2]{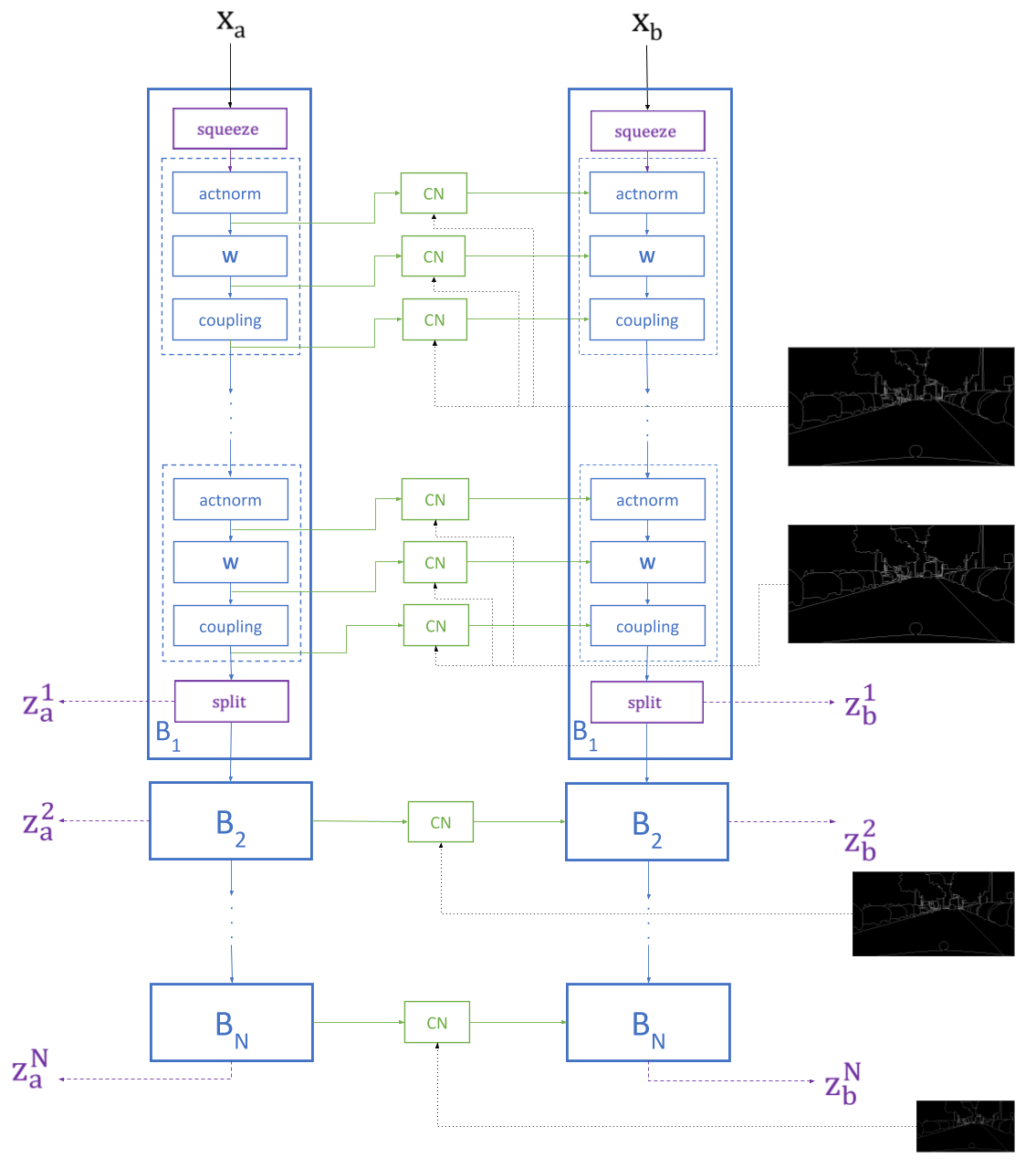}
    \caption{The proposed architecture, making all the layers conditional by inserting conditioning networks. $\mathbf{x}_{a}$ and $\mathbf{x}_{b}$ are the image pairs in the source and target domains respectively. Conditioning networks also allow for using other side information for conditioning (here shown as boundary maps, down-sampled to match spatial dimension of each Block).}
    \label{fig:extended_arch}
\end{figure}

Using boundary maps as side information enables the model to better distinguish between the objects available in the segmentation mask. In our experiments, we used boundary maps in all conditioning networks of the model by concatenating them along the channel dimension with the input to the conditioning networks \footnote{We used implementation from \cite{wang2018high} to derive the boundary maps.}. In order to do so, we down-sampled the boundary maps so they match the spatial dimension of each Block (see Figure \ref{fig:extended_arch}). We used two versions of boundary maps. In the first version, when down-sampling the boundary maps (using bi-linear interpolation), the pixel values that were previously either 0 or 1 will take float values between 0 and 1. In the second version, we forced the values to strictly be either 0 or 1, but that did not improve the performance. The results could be seen in Table \ref{tab:b_map_with_deep} (Configs. B and C of our model).

\vspace{-3mm}
\paragraph{Deepening the model.}
Gradient-checkpointing \cite{checkpointing} is an effective technique that allows fitting deeper models in GPU memory by storing activations only at specific checkpoints in the model and recompute activations between checkpoints in the backward pass (lower memory usage at the cost of longer iteration time). We used this technique to train deeper models with the same training procedure as before. A checkpoint is inserted after each Flow in the model. The deeper models have 120 Flows in total (as opposed to 64 for the shallower models). We experimented with different versions of deep models. In Config. D, the model has equal Flows in each Block. In Config. E, more Flows are concentrated in the initial Blocks (closer to data space) since these Blocks are known to be responsible for lower-level details of the synthesized image as opposed to higher Blocks, closer to latent space, that are responsible for higher-level, abstract concepts \cite{dinh2016density}. We also observed the effect of number of Blocks while keeping total number of Flows unchanged. While the number of Blocks and Flows can be seen as a choice of hyper-parameters, we observe that the optimal Blocks and Flows in this task seems to be in Config. H which has fewer Blocks with more concentrated Flows.

\DrawDeepTable

\section{Examples of effect of temperature}
\label{temperature}
This part provides visual examples of the effect of the temperature. Figures \ref{fig:temp_effect_1} and \ref{fig:temp_effect_2} show how the colors and structures change with different temperature values. We can see that samples generated with higher temperatures seem to have more vibrant colors and show more diversity, while samples generated with lower temperatures have dim colors with small variations, but the structures are generally better well-maintained.

It can be observed that synthesis quality is not as good for very low temperatures, close to 0. Such low-temperature samples are known to be atypical and nonrepresentative, as discussed and exemplified in \cite{holtzman2020curious,huszar2017gaussian} and \cite[App.\ A.1]{vandenoord2018parallel}. This is a consequence of the geometry of high-dimensional spaces, where the highest-density regions tend to contain negligible mass overall (see again \cite{huszar2017gaussian}), a notion formalized by the asymptotic equipartition property from information theory. Samples drawn at the low-temperature extreme are thus not well indicative of model quality, and flow-based models tend to use a temperature closer to 1 when generating samples in practical applications, with the optimal temperature being dependent on the model and the task.

\DrawTempFigOne
\DrawTempFigTwo

\section{Other details of the models}
\label{sec:models_details}
This part elaborates on the details of the models used in the paper.
\begin{itemize}
    \item The conditioning networks for actnorm and $1 \times 1$ convolution have three convolutional layers with 8, 4, and 2 output channels respectively, all with stride 1. The kernel size is 1 in the first layer and 3 in the next two layers.
    
    \item The first three fully connected layers in the conditioning networks have a depth of 32, 64, and 48 respectively. The depth of the last fully connected layer is determined by the number of parameters of the conditional operation. For a $c \times h \times w$ tensor, it generates $2c$ parameters for the actnorm operation and $c^2$ parameters for the $1 \times 1$ operation.
    
    \item The conditioning network for the coupling layer has two convolutional layers, both of which have 128 output channels with kernel size of 3 and stride of 1.
    
    \item The first version of C-Glow \cite{lu2020structured} has 3 Blocks each having 8 Flows. The number of filters in the convolutional layer of the coupling operation is 104, the convolutional layer of the conditioning network has 52 channels, and the width of the linear layer in the conditioning network is 26. These numbers are chosen to reflect the same ratio as the original numbers used by the authors while being feasible to fit in GPU memory.
    
    \item The second version of C-Glow \cite{lu2020structured} has 4 Blocks each having 16 Flows. The number of channels of the coupling layer, convolutional layer of condition networks and the width of the linear layer of the conditioning network are 512, 20, and 10 respectively. These were chosen so that its Glow is as similar as possible to our model and \hk{Dual-Glow} (in terms of depth) while keeping the conditioning networks as deep as possible.
    
    \item Both \hk{Dual-Glow} \cite{journals/corr/abs-1908-08074} and pix2pix \cite{Isola_2017_CVPR} were trained using the default hyper-parameters specified in their official implementations.
    
    \item For synthesizing higher-resolution images, we trained a model with 4 Blocks each with 10 Flows for 100 epochs. In the first 50 epochs we used a learning rate of $10^{-4}$. Next, we linearly decreased the learning rate to 0 in the remaining 50 epochs.
\end{itemize}

\section{Loss curve}
\label{sec:loss_curve}
\begin{figure}
    \centering
    \includegraphics[width=0.9\columnwidth]{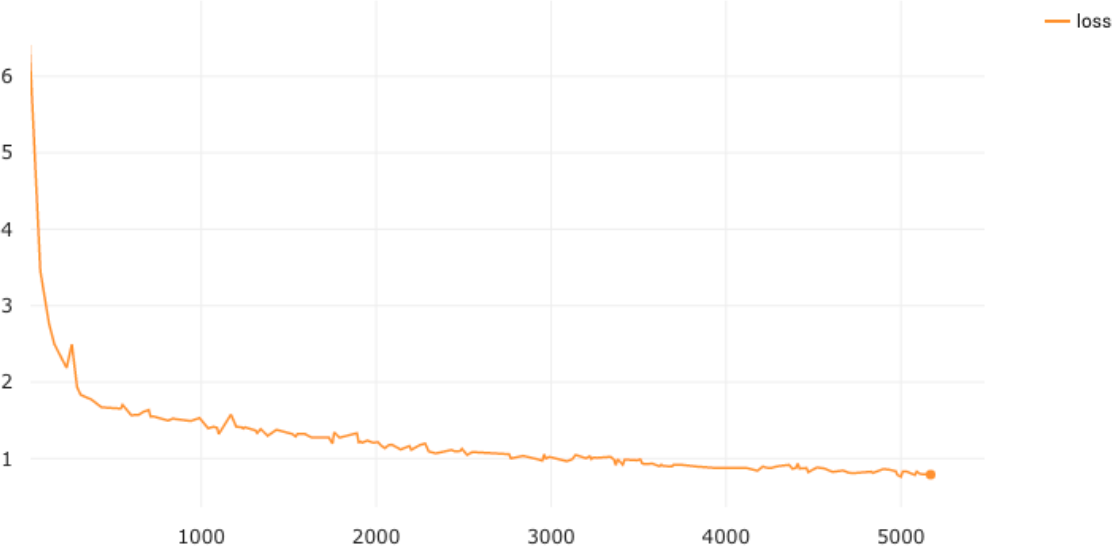}
    \caption{Example Full-Glow loss curve demonstrating stable training.}
    \label{fig:loss}
\end{figure}

By initializing all parameters using the same principles described in the original Glow paper \cite{kingma2018glow}, we found Full-Glow training to be very stable, as illustrated in Figure \ref{fig:loss}. 
In contrast, non-smooth training behavior was evident when training the pix2pix benchmark using the official code. This is consistent with mathematical findings that widely-used GAN setups such as WGAN need not converge since monotonicity is not assured \cite{mescheder2018training}.

\end{document}